\documentclass[conference]{IEEEtran}
\IEEEoverridecommandlockouts
\usepackage{cite}
\usepackage{amsmath,amssymb,amsfonts}
\usepackage{algorithmic}
\usepackage{graphicx}
\usepackage{textcomp}
\usepackage{xcolor}
\usepackage{multirow,array}
\usepackage{bm}
\usepackage{wrapfig}
\usepackage{hyperref}

\newcommand{\X}{\ensuremath{\mathbf{X}}}
\newcommand{\B}{\ensuremath{\mathbf{B}}}

\newcommand{\hh}{\ensuremath{\mathbf{h}}}
\newcommand{\x}{\ensuremath{\mathbf{x}}}
\newcommand{\aalpha}{\ensuremath{\bm{\alpha}}}

\newcommand{\y}{\ensuremath{\mathbf{y}}}

\newcommand{\Y}{\ensuremath{\mathbf{Y}}}

\newcommand{\btheta}{\ensuremath{\boldsymbol{\theta}}}

\newcommand{\f}{\ensuremath{\mathbf{f}}}

\newcommand{\bbR}{\ensuremath{\mathbb{R}}}
\newcommand{\norm}[1]{\left\lVert#1\right\rVert}

\DeclareMathOperator*{\argmin}{argmin}

\def\BibTeX{{\rm B\kern-.05em{\sc i\kern-.025em b}\kern-.08em
    T\kern-.1667em\lower.7ex\hbox{E}\kern-.125emX}}
\begin{document}

\title{Estimating Vector Fields from Noisy Time Series\\
\thanks{This work was performed under the auspices of the U.S. Department of Energy by Lawrence Livermore National Laboratory under Contract DE-AC52-07NA27344 and was supported by the LLNL-LDRD Program under Project No. 19-ERD-009. LLNL-CONF-800042.  H. S. Bhat and M. Reeves acknowledge partial support from NSF DMS-1723272. We also acknowledge use of the MERCED computational cluster, funded by NSF award ACI-1429783. {\tiny \copyright 2020 IEEE.  Personal use of this material is permitted.  Permission from IEEE must be obtained for all other uses, in any current or future media, including reprinting/republishing this material for advertising or promotional purposes, creating new collective works, for resale or redistribution to servers or lists, or reuse of any copyrighted component of this work in other works.}}
}

\author{\IEEEauthorblockN{Harish~S. Bhat}
\IEEEauthorblockA{\textit{Applied Mathematics} \\
\textit{University of California, Merced}\\
Merced, CA USA \\
hbhat@ucmerced.edu}
\and
\IEEEauthorblockN{Majerle Reeves}
\IEEEauthorblockA{\textit{Applied Mathematics} \\
\textit{University of California, Merced}\\
Merced, CA USA \\
mreeves3@ucmerced.edu}
\and
\IEEEauthorblockN{Ramin Raziperchikolaei}
\IEEEauthorblockA{
\textit{Rakuten, Inc.}\\
San Mateo, CA USA \\
ramin.raziperchikola@rakuten.com}
}

\maketitle

\begin{abstract}
While there has been a surge of recent interest in learning differential equation models from time series, methods in this area typically cannot cope with highly noisy data.  We break this problem into two parts: (i) approximating the unknown vector field (or right-hand side) of the differential equation, and (ii) dealing with noise.  To deal with (i), we describe a neural network architecture consisting of tensor products of one-dimensional neural shape functions. For (ii), we propose an alternating minimization scheme that switches between vector field training and filtering steps, together with multiple trajectories of training data.  We find that the neural shape function architecture retains the approximation properties of dense neural networks, enables effective computation of vector field error, and allows for graphical interpretability, all for data/systems in any finite dimension $d$.  We also study the combination of either our neural shape function method or existing differential equation learning methods with alternating minimization and multiple trajectories.  We find that retrofitting any learning method in this way boosts the method's robustness to noise.  While in their raw form the methods struggle with 1\% Gaussian noise, after retrofitting, they learn accurate vector fields from data with 10\% Gaussian noise.
\end{abstract}

\section{Introduction}
\label{s:prob}
We consider the problem of learning a dynamical system from multiple, vector-valued time series.  Suppose we have $N$ time series or trajectories, each observed at $T$ discrete times $\{t_i\}_{i=1}^T$.  We assume this temporal grid is sufficiently fine to capture the dynamics of the system of interest.  Let $\y^j_i \in \mathbb{R}^d$ denote the observation for trajectory $j$ at time $t_i$.  Here $d \geq 1$ is arbitrary.  Given this data, we compute estimates of (i) the states $\widehat{\x}^j_i$ and (ii) a vector field $\widehat{\f} : \bbR^d \to \bbR^d$ that determines a dynamical system model for the time-evolution of the states.

Suppose the true trajectory $\x^j(t)$ satisfies the nonlinear system of differential equations given by
\begin{equation}
	\label{eq:ode}
	\dot\x^j(t) = {d\x^j(t)}/{dt} = \f(\x^j(t)).
\end{equation}
We model the observation $\y^j_i$ as the true state plus noise:
\begin{equation}
\label{eq:noise}
\y^j_i = \x^j(t_i) + \epsilon^j(t_i), \quad i=1,\dots,T.
\end{equation}
We seek $\widehat{\x}^j_i$ that approximates $\x^j(t_i)$, and $\widehat{\f}$ that approximates $\f$.  While there has been much recent interest in learning differential equations from data, the bulk of the literature focuses on computing $\widehat{\f}$ from relatively clean data; \cite{Schaeffe17} notes that when Gaussian noise of strength greater than $3 \%$ is present in observed states, estimation becomes unstable and inaccurate.  We demonstrate below that leading methods encounter difficulty even at relatively low noise magnitudes \cite{Brunton16, Raissi17}. Our method yields accurate estimates even when the data is corrupted by $10\%$ noise.

Our prior work \cite{RaziBhatICML2019} established that the block coordinate descent proximal filtering method yields more accurate and robust parameter estimates than either iPDA or the extended Kalman filter.  A key element of our filtering approach is its proximal step \cite{Parikh14}.  In contrast, other techniques such \emph{iPDA} and \emph{soft adherence} use a penalty term that anchors the filtered states (at all iterations) to the data $\y$ \cite{RamsayHooker, Rudy19}.  Our work shares goals with \cite{tran_exact_2017}, who are concerned with recovering the functional form of sufficiently ergodic, chaotic dynamical systems from highly corrupted measurements.  While our parameterizations of $\f$ differ considerably, both the present paper and \cite{tran_exact_2017} develop and apply alternating minimization methods.  

The present paper focuses on extending \cite{RaziBhatICML2019} to the setting where the functional form of $\widehat{\f}$ is unknown and must be estimated. In our prior work, we assumed this vector field was known up to a finite-dimensional set of parameters.  Hence our prior work contains no mention of neural networks, nor does it contain comparisons against the methods of \cite{Raissi17} and \cite{Rudy18a}.

The central finding of this paper is as follows: as the magnitude of noise increases, computing accurate $\widehat{\x}^j_i$ and $\widehat{\f}$ is still possible if one alternates model estimation steps with filtering steps, and if one trains using a larger number $N$ of trajectories.   Using tests on simulated data, we show that leading equation discovery methods can be retrofitted with our filtering approach, greatly enhancing the ability of these methods to cope with noisy data.  These methods differ in the way they model the estimated vector field $\f$: sparse linear combinations of prescribed functions \cite{Brunton16}, neural shape functions, or dense, feedforward neural networks \cite{Raissi17}.  We conduct these tests for both the  FitzHugh--Nagumo system and a nonlinear oscillator chain, showing that we can recover both the filtered states and the underlying vector field with high accuracy.

We also apply our method to power grid data recorded by micro-phasor measurement units ($\mu$PMUs).  The units record synchronized measurements of voltage and phase angle at the power distribution level \cite{Cadena19}.  Using data taken on two different days for which the system's status has been labeled, we estimate two vector fields, one for normal operation and one for anomalous operation.  To estimate these vector fields, we found it essential to both filter the data and increase the number of trajectories.

\section{Methods}
\subsection{Filtering}
\label{sect:opt}
Let $\widehat{\f}(\x, \btheta)$ model the true vector field $\f$.  For now, we are not concerned with the actual details of this model except to say that the model parameters are given by $\btheta$.  Consider an explicit Euler time discretization of (\ref{eq:ode}) with time step $\Delta_i = t_{i+1} - t_i$.  For trajectory $j$, we have
\begin{equation}
\label{eq:ode_em}
    \x^j_{i+1} - \x^j_i = \f(\x^j_i,\btheta) \Delta_{i}, \quad i=1,...,T-1
\end{equation}
We use Euler purely for simplicity here; in practice the method can accommodate higher-order, explicit time integration schemes.  Let $\widehat{\X}$ denote the collection $\{\widehat{\x}^j_i\}_{j=1, i=1}^{N, T}$ of all filtered states over all trajectories.  Then define the objective function
\begin{equation}
\label{eq:objx}
    E(\widehat{\X},\widehat{\btheta}) = \sum_{j=1}^N {\sum_{i=1}^{T-1}{ {\norm{\frac{\widehat{\x}^j_{i+1} - \widehat{\x}^j_{i}}{\Delta_{i}}  - \widehat{\f}(\widehat{\x}_i^j,\widehat{\btheta})}}^2} }.
\end{equation}
Assume that if the true states $\X$ were known, that $\widehat{\f}$ could be trained by minimizing $E$ over $\btheta$.  We thus refer to minimization over $\btheta$ as model training.

Let $\Y$ denote the collection $\{\y^j_i\}_{j=1, i=1}^{N, T}$ of all observations over all trajectories.  We propose the following alternating procedure to learn $\widehat{\X}$ and $\widehat{\btheta}$.  The states are initialized to be the data: $\widehat{\X}^0 = \Y$---superscripts denote the iteration number:
\begin{subequations}
	\label{eq:multistepiter}
	\begin{align}
	\label{eq:multisteptheta}
		\text{train: } \widehat{\btheta}^{k+1} &= \argmin_{\btheta} E(\widehat{\X}^{k},\btheta) \\
	\label{eq:multistepX}
		\text{filter: } \widehat{\X}^{k+1} &= \argmin_{\X} \left\{ E(\X, \widehat{\btheta}^{k+1}) + \lambda \| \X - \widehat{\X}^{k} \|^2 \right\}.
	\end{align}
\end{subequations}
We terminate when the change in $(\widehat{\btheta},\widehat{\X})$ is sufficiently small.

\subsection{Three Parameterizations of $\widehat{\f}$}
\label{sect:parameterize}

\subsubsection{SINDy} We first take $\widehat{\f}$ to be a linear combination of prescribed functions.  Given a $1 \times d$ input $\x$ we let $\Xi(\x)$ denote a $1 \times s$ dictionary of functions.  For instance, for $(x_1, x_2)$ we can take $\Xi(x_1, x_2) = (x_1, x_2, x_1^2, x_2^2, x_1 x_2)$, i.e., all polynomials in the components of $x$, up to degree two.  Using this dictionary, we write
\begin{equation}
	\label{eqn:sindy}
    \widehat{\f}(\x) = \Xi(\x) \btheta
\end{equation}
where $\btheta$ is an $s \times d$ matrix of coefficients. Throughout this paper, to extend $\widehat{\f}(\x)$ to a function $\widehat{\f}(\X)$, we apply $\widehat{\f}$ to each $1 \times 1 \times d$ slice of $\X$.  To solve for $\btheta$, we apply an iteratively thresholded least-squares regression procedure that promotes a sparse solution $\btheta$---for further details, consult \cite{Brunton16, ZhangSchaeffer2019}.  

\subsubsection{Neural Shape Functions} Instead of prescribing $\Xi$ and learning only $\widehat{\btheta}$, here we learn both $\Xi$ and $\widehat{\btheta}$.  Let $B$ be the desired number of one-dimensional shape functions; we model these using a dense neural network with one-dimensional input, $D-1$ hidden layers, and a final layer with $B$ outputs. Let $\hh(x) \in \mathbb{R}^B$ denote the output corresponding to scalar input $x$; then $h_{j}$ is the $j$-th one-dimensional shape function.

\begin{figure}[t] 
	\begin{center}
	\begin{tabular}{c@{}c@{}c@{}}
		& {{pred. err on $[0, T]$}} & \\
				$1\%$ noise & $5\%$ noise & $10\%$ noise \\
		\includegraphics*[width=0.3\linewidth,trim={0 0 0 .1ex},clip]{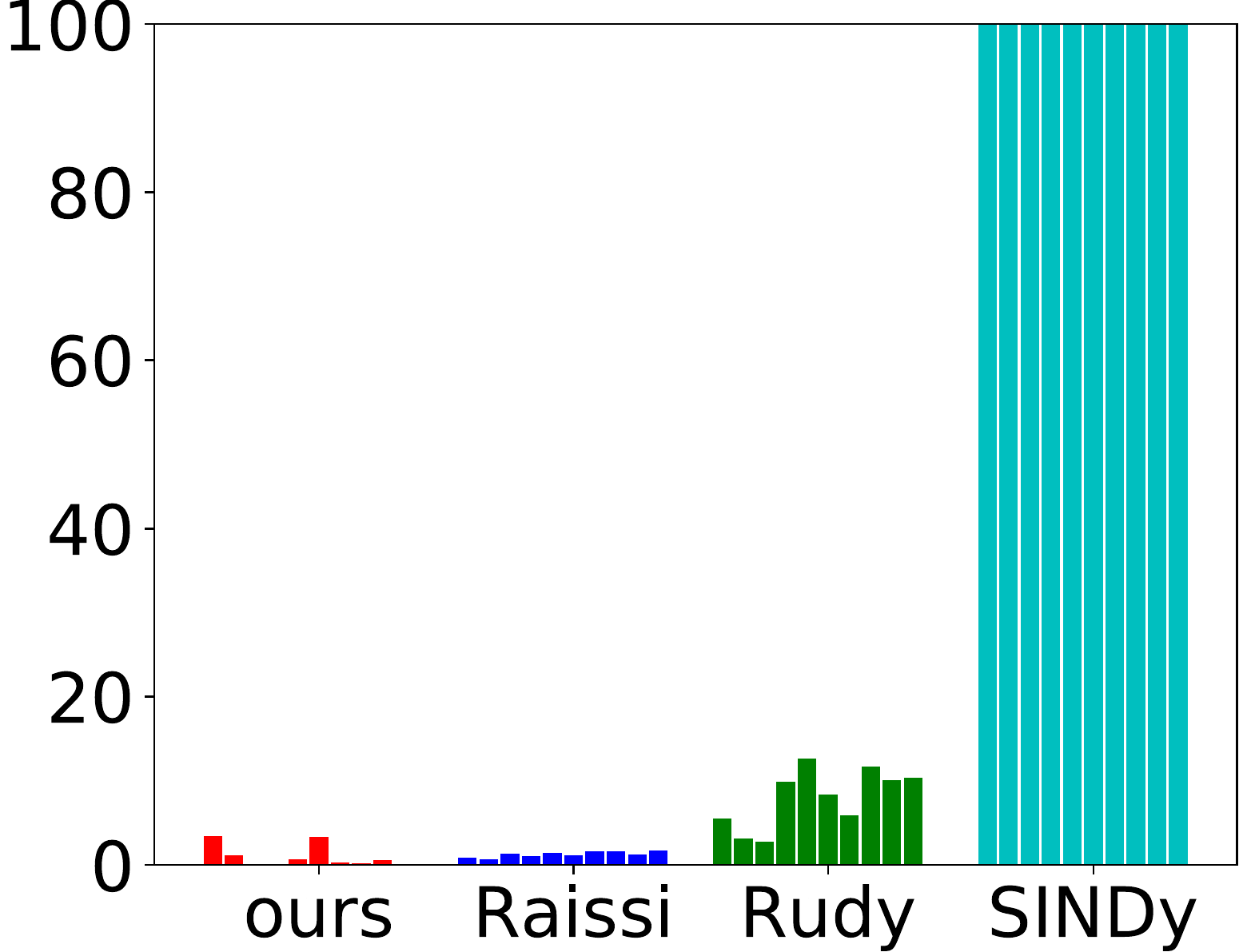} &
		\includegraphics*[width=0.3\linewidth,trim={0 0 0 .1ex},clip]{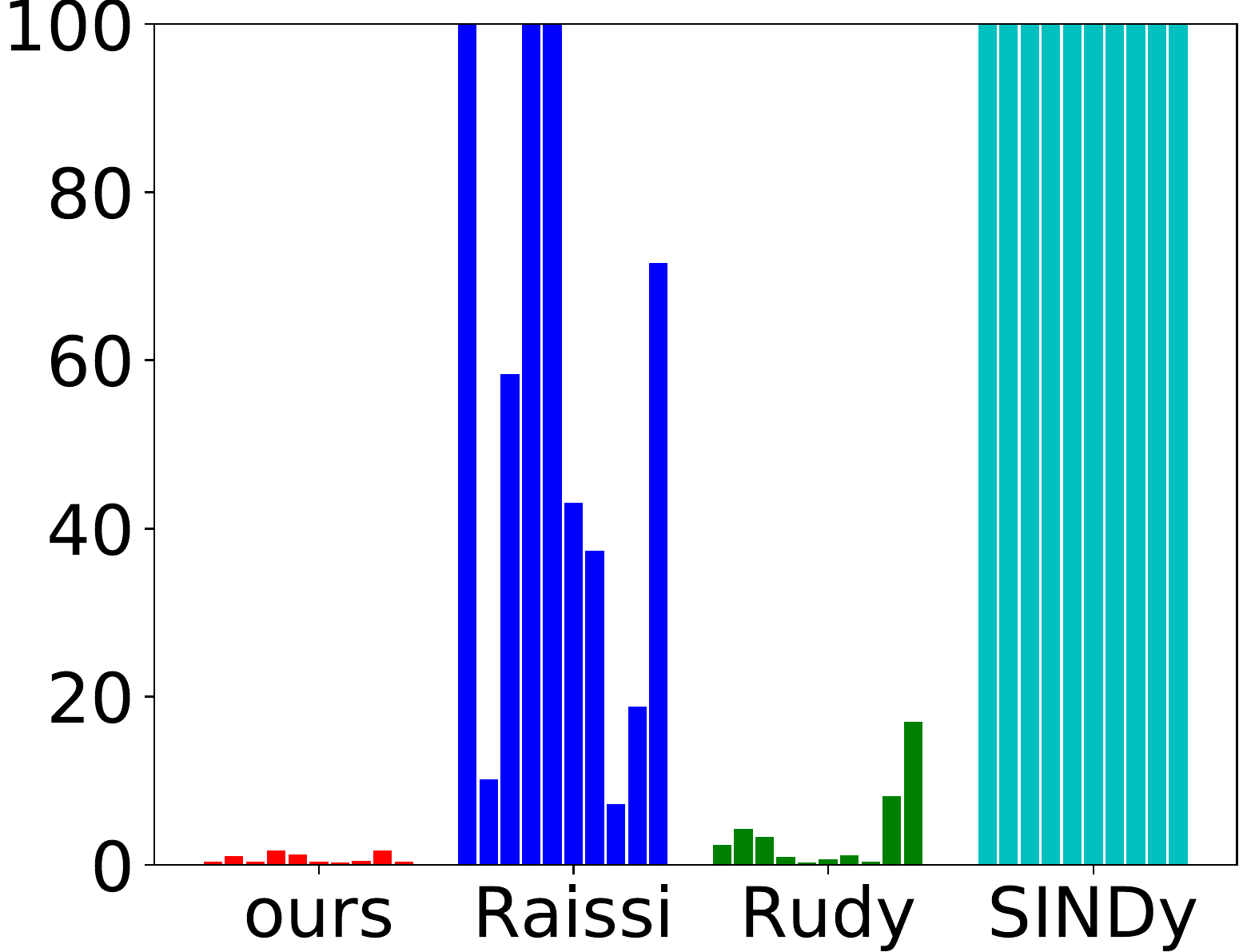} &
		\includegraphics*[width=0.3\linewidth,trim={0 0 0 .1ex},clip]{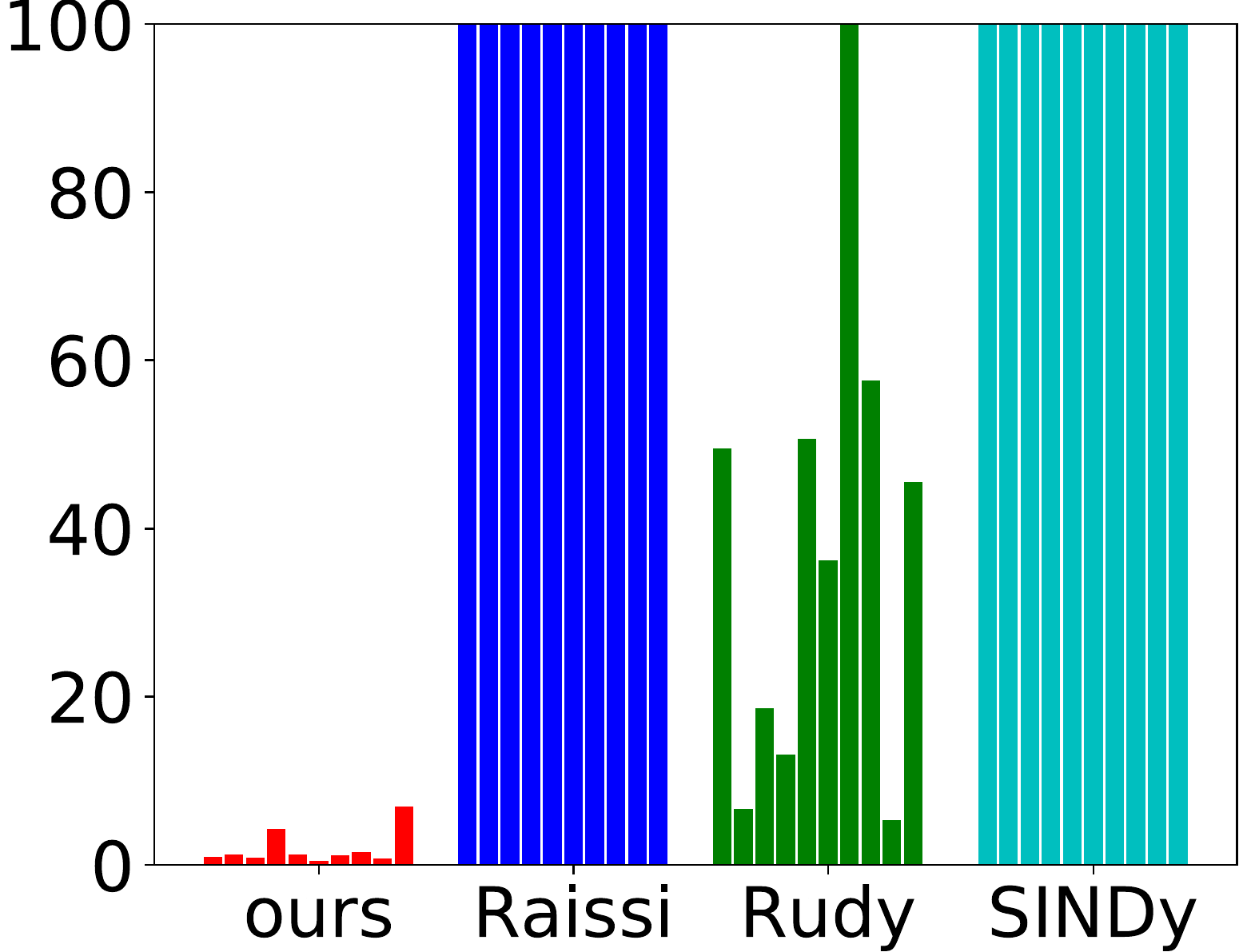} \\
	\end{tabular}
	\begin{tabular}{c@{}c@{}c@{}}
		& {{pred. err on $[T, 2T]$}} & \\
						$1\%$ noise & $5\%$ noise & $10\%$ noise \\
                \includegraphics*[width=0.3\linewidth,trim={0 0 0 .1ex},clip]{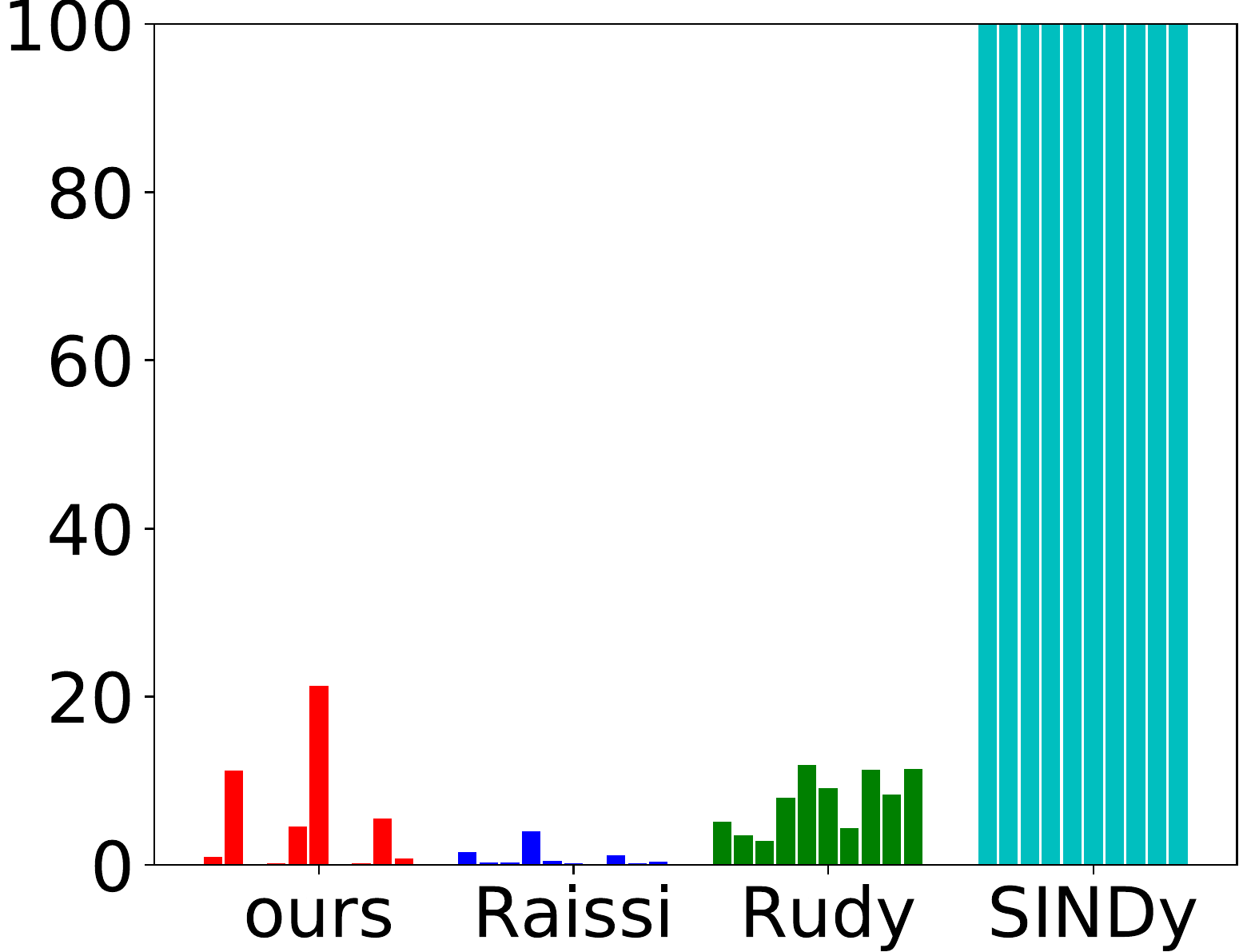}&
                \includegraphics*[width=0.3\linewidth,trim={0 0 0 .1ex},clip]{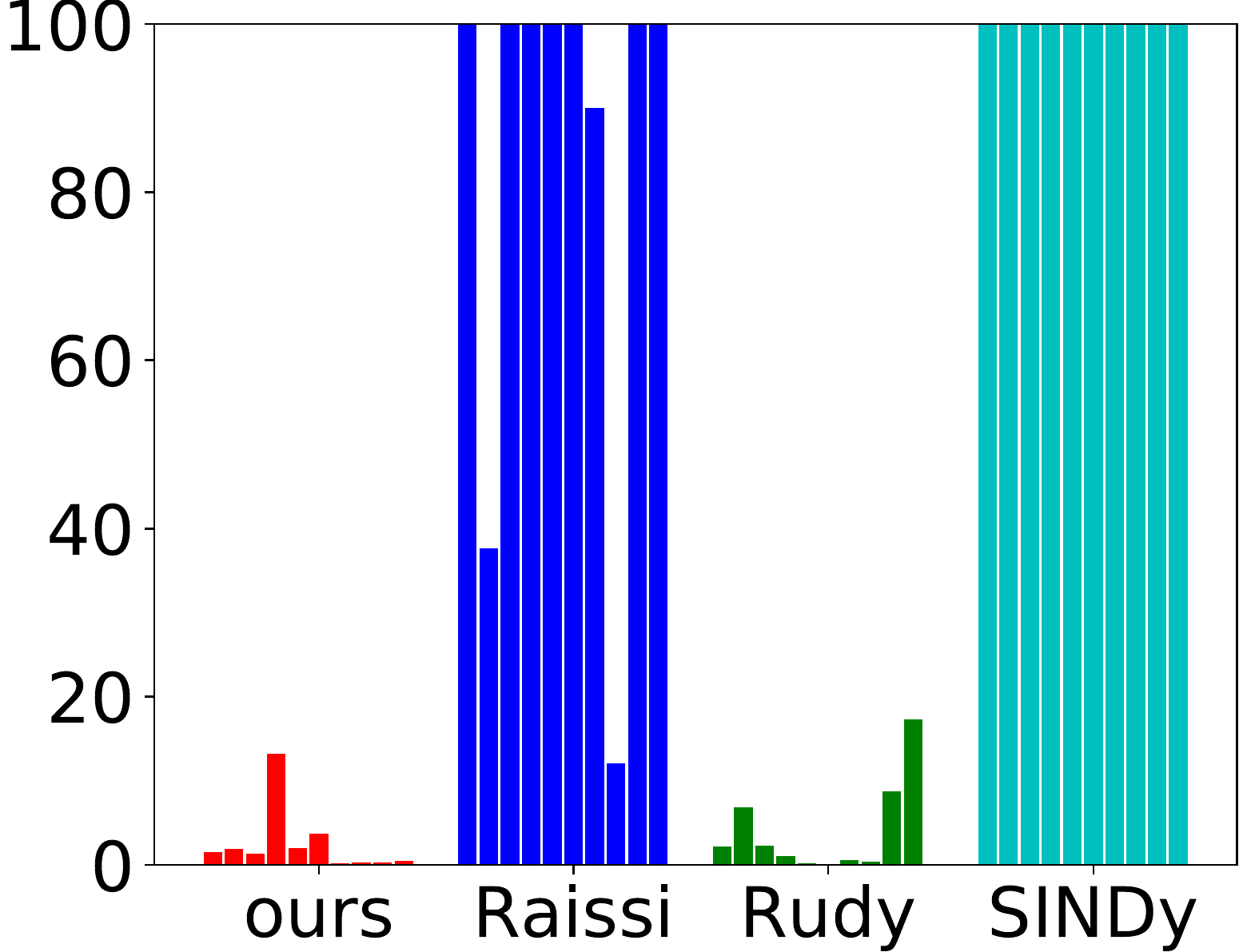}&
                \includegraphics*[width=0.3\linewidth,trim={0 0 0 .1ex},clip]{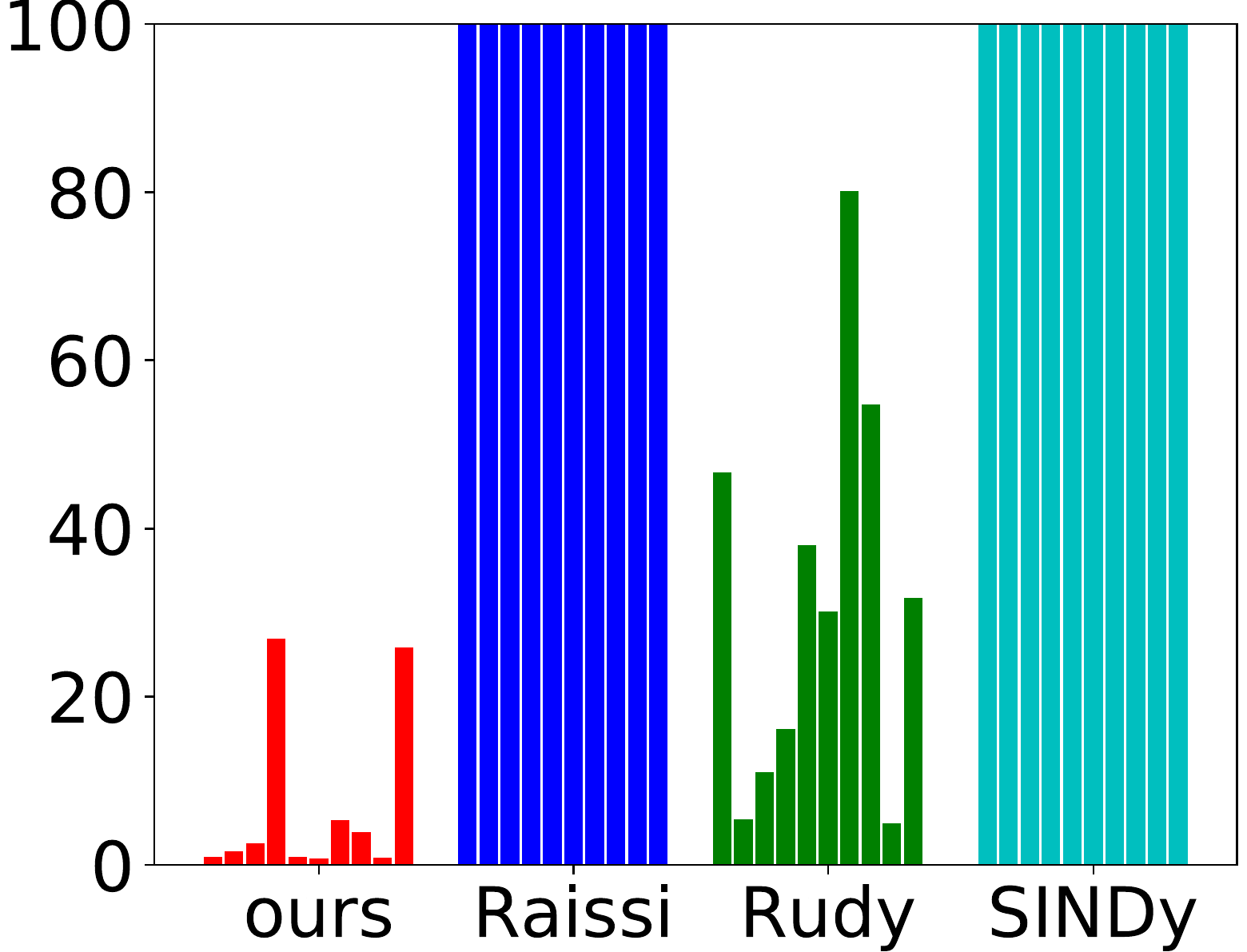}\\
	\end{tabular}
	\end{center} \vspace{-0.5cm}
	\caption{We compare the neural shape method with \cite{Raissi17}, \cite{Rudy18a}, and \cite{Brunton16} on FitzHugh--Nagumo with different noise levels.  For each noise level, we create $10$ sets of observations, run the methods on each set separately, and report prediction errors. Each error bar shows the error on one experiment.  For all plots, we train on the time interval $[0,T]$.  In the upper (resp., lower) set of plots, we report the prediction error on $[0,T]$ (resp., $[T, 2T]$).} 
	\label{f:comp}
\end{figure}

To create multi-dimensional shape functions, we take tensor products of one-dimensional shape functions.  Let a multi-index denote a collection of integers $\aalpha = (\alpha_1, \ldots, \alpha_d)$ such that $0 \leq \alpha_k \leq B$.  Define $h_0(x) = 1$.  Then for a given multi-index $\aalpha$, define the multi-dimensional shape function
\begin{equation}
\label{eq:multd}
	H_{\aalpha}(\x) = \prod_{k=1}^d h_{\alpha_k} (x_k) \in \mathbb{R}.
\end{equation}
Let $A$ be a collection of multi-indices and $|A|$ its cardinality.  Suppose we compute $H_{\aalpha}(\x)$ for each $\aalpha \in A$; in this way we obtain a $1 \times |A|$ vector $\mathbf{H}(\x)$.  Here $\mathbf{H}(\x)$ plays the same role as $\Xi(\x)$ in SINDy above; $|A|$ is analogous to $s$, the number of shape functions.  Now let $\B$ denote an $|A| \times d$ matrix of weights.  The neural shape function model is 
$\widehat{\f}(\x) = \mathbf{H}(\x) B$.  Note that there are $(B+1)^d$ possible multi-dimensional shape functions.  In practice, we choose $A$ such that $|A| \ll (B+1)^d$, thus constraining $\widehat{\f}$ to be a small linear combination of multi-dimensional shape functions. 
 The set of parameters $\btheta$ consists of $B$ together with all weights and biases in the $\hh(x)$ network.
  
Suppose we work in a compact subset $\mathcal{K} \subset \mathbb{R}^d$.  Given direct observations of a smooth vector field $\f$ in $\mathcal{K}$, it is possible to achieve arbitrarily small error $\| \f - \widehat{\f} \|$ by choosing the neural shape function hyperparameters sufficiently large. 
 To understand why, note that universal approximation theory guarantees that (even when $D=1$) the space of our one-dimensional neural shape functions is dense in the space of continuous functions.  Hence tensor products of these shape functions are dense in the space of tensor products of continuous functions.  By Stone-Weierstrass, we see that any smooth vector field can be approximated by linear combinations of tensor products of continuous functions (e.g., univariate polynomials).  Putting the previous two facts together, we conclude that linear combinations of our multi-dimensional neural shape functions can be used to approximate smooth vector fields; the accuracy of this approximation can be controlled by the number of units in each $h_j$ network together with $|A|$.
   
\subsubsection{Dense Neural Network (DNN)} We also consider a dense, feedforward neural network model $\widehat{\f}(\x)$ with $d$ inputs and $d$ outputs, the model used by \cite{Raissi17}.

\section{Results}
Here we present results from simulated data experiments in which the ground truth vector field $\f$ is known. 

\subsection{FitzHugh--Nagumo}

\begin{figure}
\centering
	\includegraphics[width=0.3\textwidth,trim={2.5cm 2cm 2cm 2.5cm},clip]{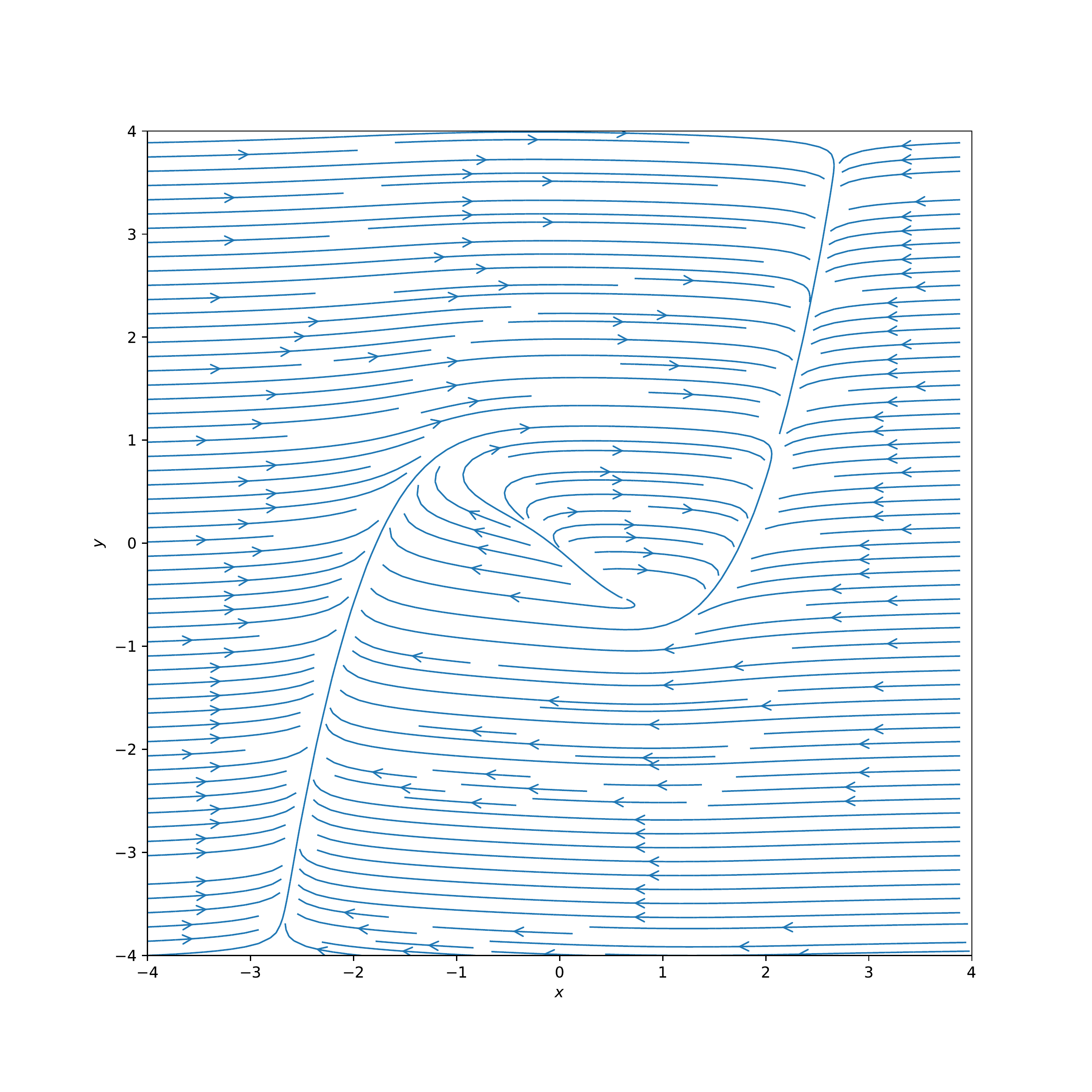} \vspace{-0.5cm}
	\caption{Ground truth FitzHugh--Nagumo vector field.}
	\label{fig:fitztruth}
\end{figure}

In Figure \ref{f:comp}, we compare neural shape functions plus filtering (our method) against three published methods \cite{Rudy18a,Raissi17,Brunton16} on simulated data from the FitzHugh--Nagumo system:
\[
	\frac{dx_0}{dt} = c \left(x_0 - \frac{x_0^3}{3} + x_1\right), \qquad
	\frac{dx_1}{dt} = - \frac{1}{c} (x_0 - a + b x_1).
\]
To generate simulated data, we set the system parameters $a=0.5$, $b=0.2$, $c=3$ and initial condition $\x_{1}=[-1,1]$.  We then numerically solve the system on the interval $[0,T]$ with $T=20$, recording data at a spacing of $\Delta t = 0.05$.  To these clean trajectories we add mean-zero Gaussian noise at strengths of $1\%$, $5\%$, and $10\%$.  We train using data on $[0,T]$; in Figure \ref{f:comp}, we plot prediction errors for both the training interval and an extrapolatory test interval $[T, 2T]$. 

For this initial comparison, let us define the concept of prediction error.  Suppose we are at iteration $k$ and that we have trained the model for $e$ epochs. Define the \emph{predicted states} $\widetilde{\X}^k_e$ to be the numerical solution of (\ref{eq:ode}) that results from using the filtered initial conditions contained in $\widehat{\X}^k$ (for the $j$-th trajectory, $\widehat{\x}^j_1$) together with the current best vector field $\widehat{\f}(\x, \widehat{\btheta}^k)$.  We call the distance between $\widetilde{\X}^{k}_e$ and $\X^{k}$ the \emph{prediction error}---we claim this is a good metric to detect overfitting. If we find that the current prediction error increases as we increase the number of epochs, we halt the optimization step (\ref{eq:multisteptheta}), set $\widehat{\btheta}^k$ to the weights of the network with minimum current prediction error, and proceed to the filtering step (\ref{eq:multistepX}).

\begin{figure}[!t]
  \centering
  \begin{tabular}{@{}c@{\hspace{1ex}}c@{}c@{}}
  & dimension $j = 0$ & dimension $j = 1$ \\[-.2ex]
  	\rotatebox{90}{\hspace{4ex} \small{$f_j(\x)$}} &       
    \includegraphics*[width=0.45\linewidth,trim={0 0 0 4ex},clip]{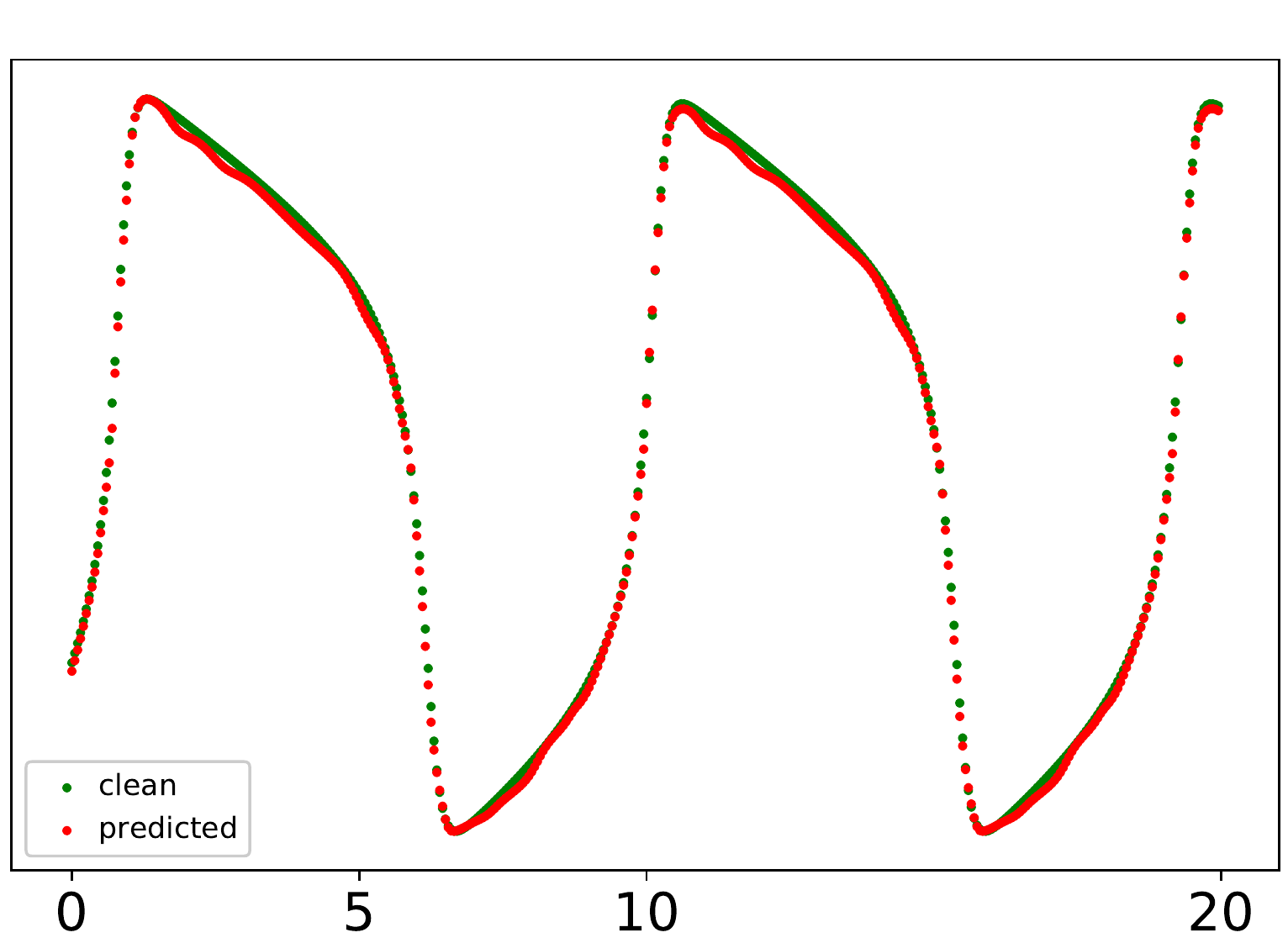}&
	\includegraphics*[width=0.45\linewidth,trim={0 0 0 4ex},clip]{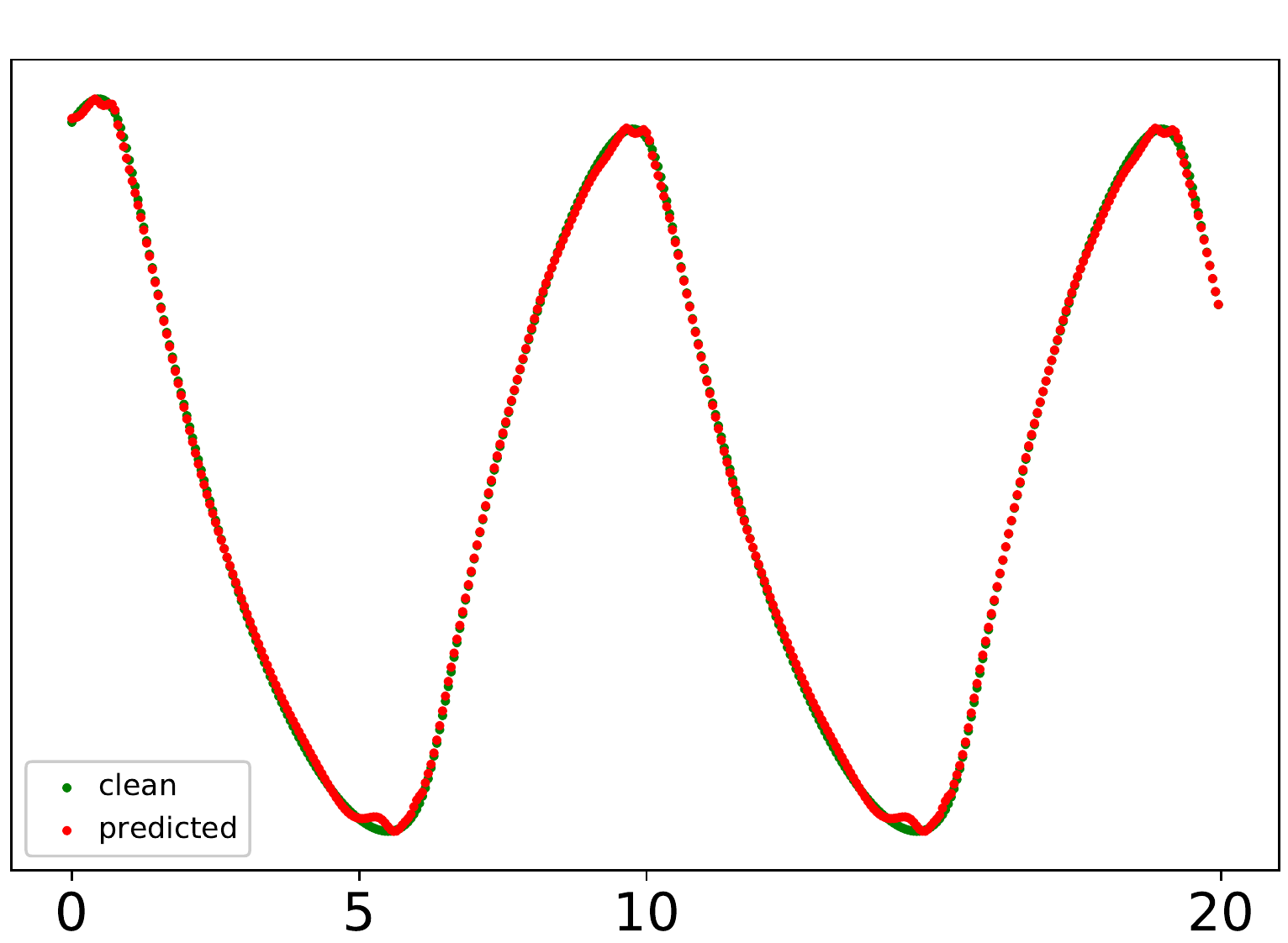}\\[-1ex]
   	& time $t$ & time $t$\\[2ex]
  \end{tabular}
   \begin{tabular}{@{}c@{\hspace{.5ex}}c@{}c@{}c@{}}
     & shape $1$: $h_1(x_k)$ & shape $2$: $h_2(x_k)$ & shape $3$: $h_3(x_k)$\\[-.1ex]
   	\rotatebox{90}{\hspace{2ex} \small{$h(x_k)$}} &       
	 \includegraphics*[width=0.3\linewidth,trim={0 0 0 4ex},clip]{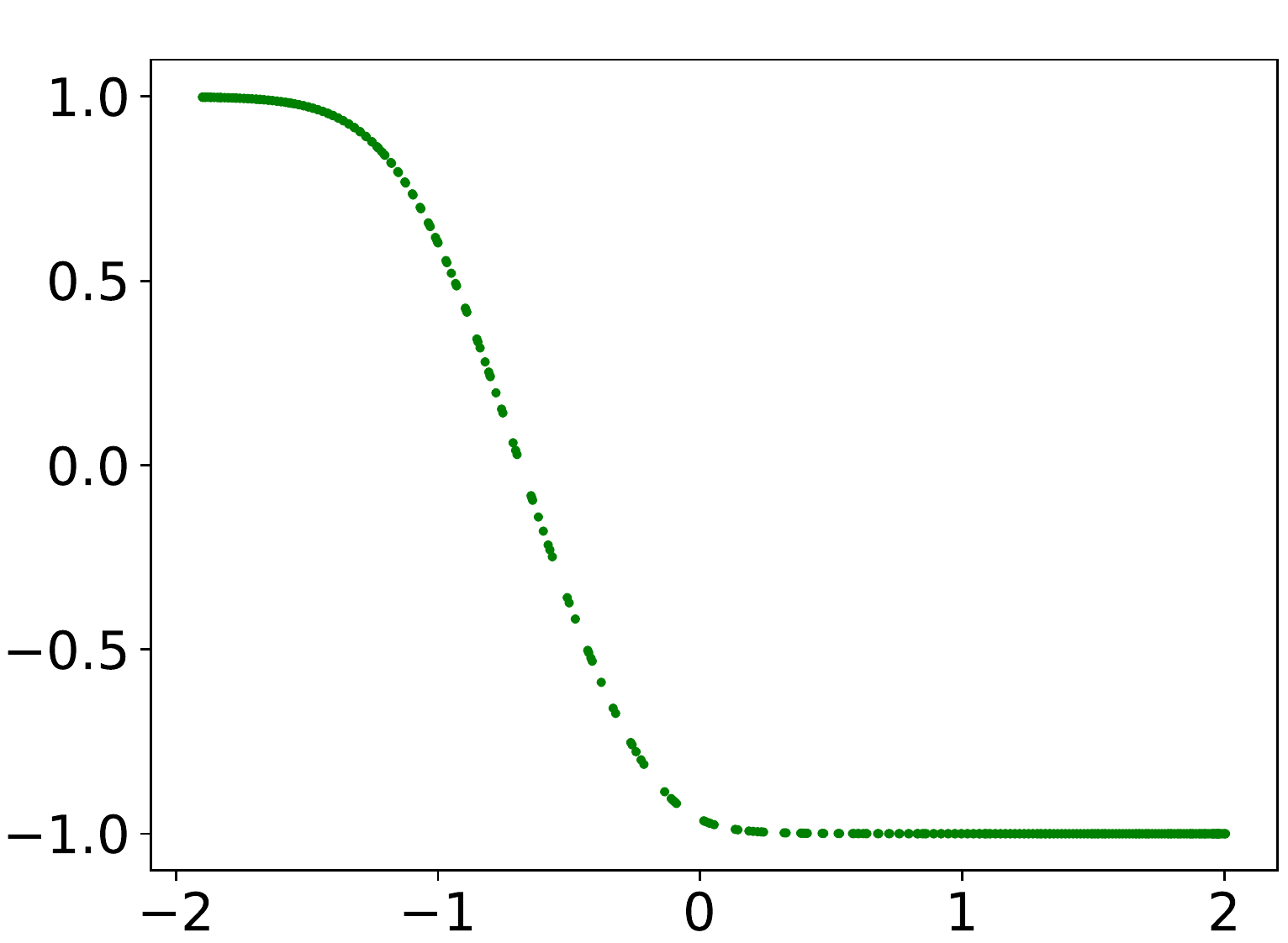}&
	 \includegraphics*[width=0.3\linewidth,trim={0 0 0 4ex},clip]{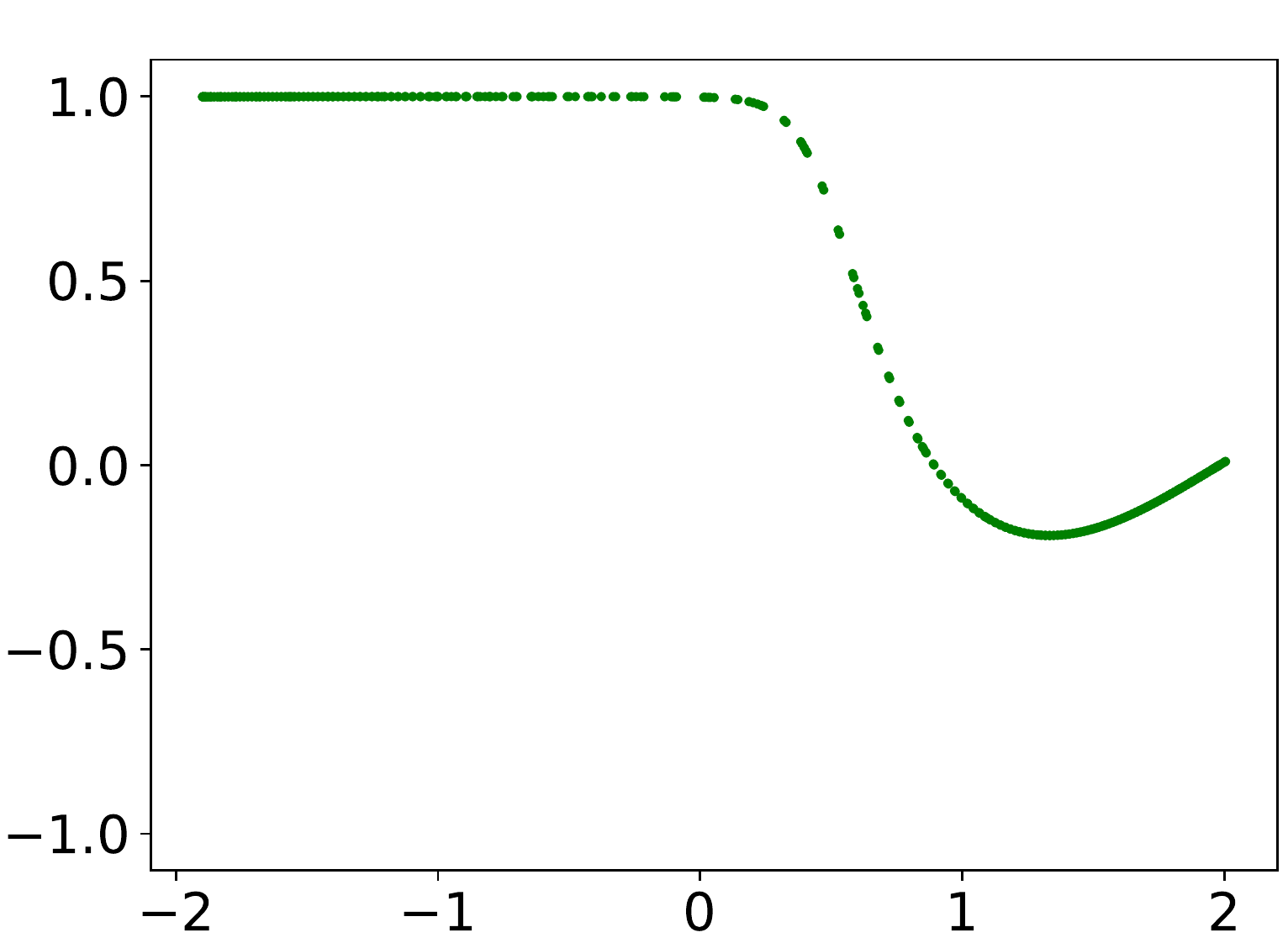}&
	 \includegraphics*[width=0.3\linewidth,trim={0 0 0 4ex},clip]{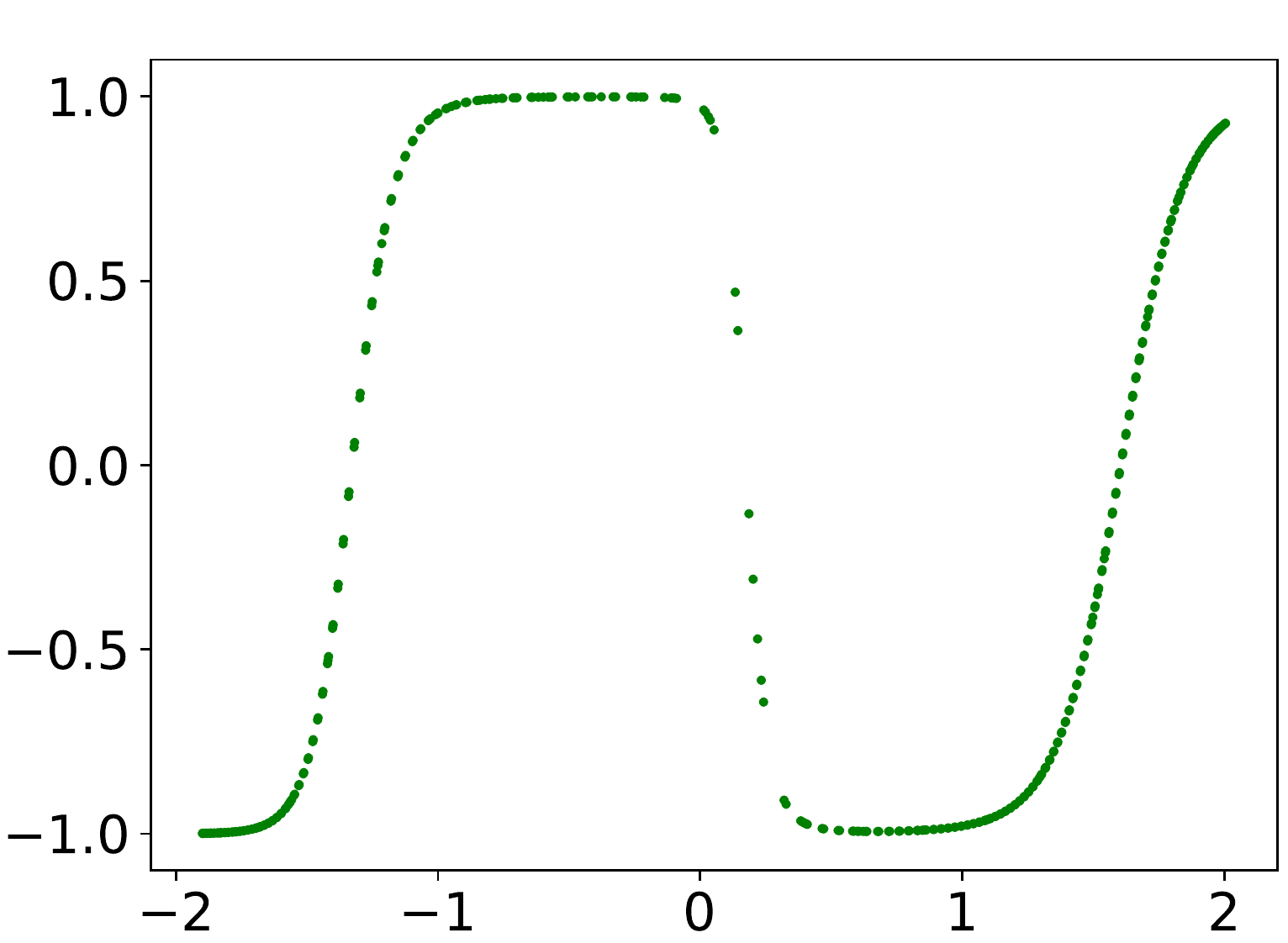}\\[-1.2ex]
	  & $x_k$ & $x_k$ & $x_k$\\[2ex]
	  & shape $4$: $h_4(x_k)$ & shape $5$: $h_5(x_k)$ & shape $6$: $h_6(x_k)$\\[-.1ex]
   	\rotatebox{90}{\hspace{2ex} \small{$h(x_k)$}} &       
	  \includegraphics*[width=0.3\linewidth,trim={0 0 0 4ex},clip]{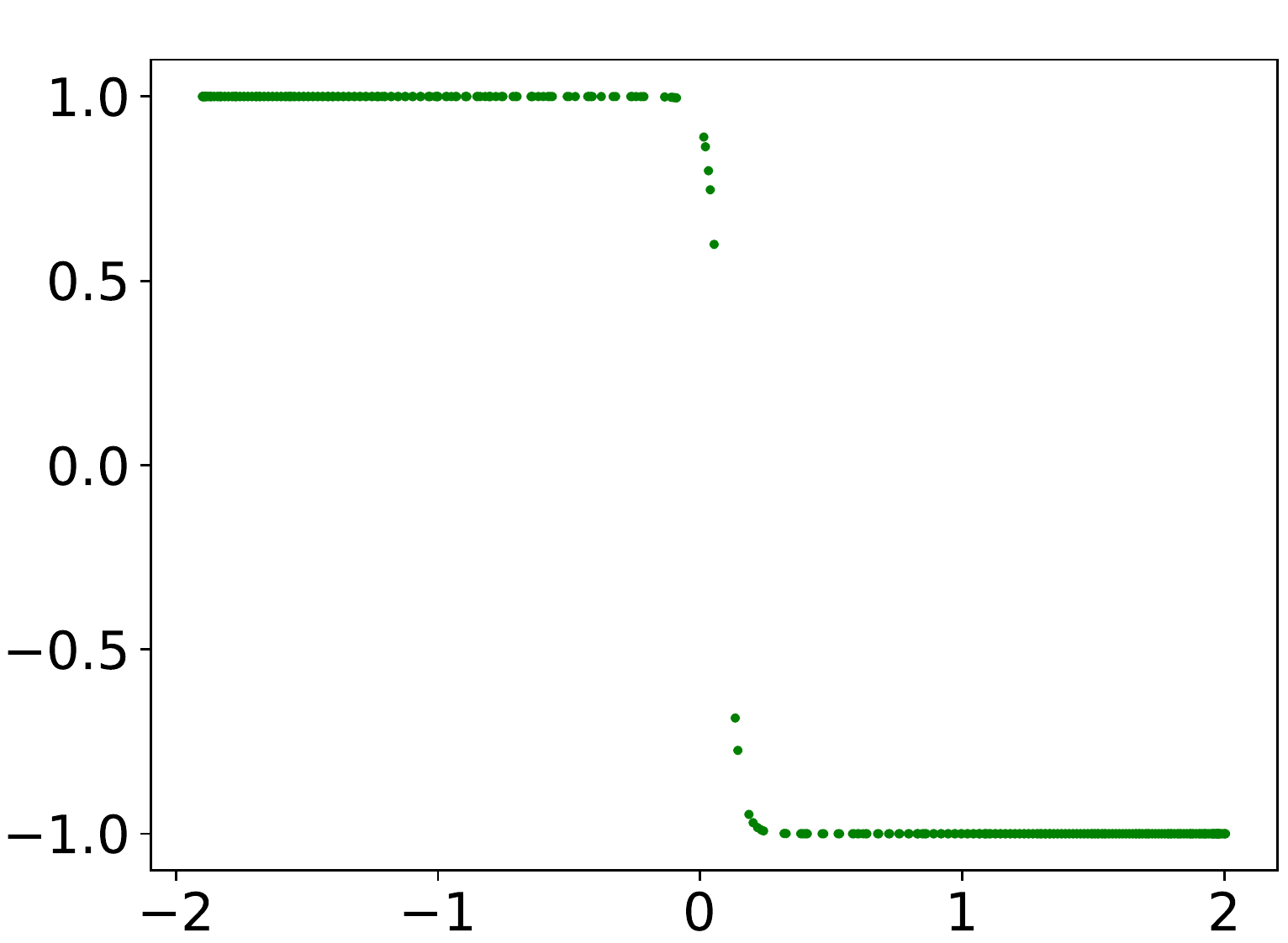}&
	 \includegraphics*[width=0.3\linewidth,trim={0 0 0 4ex},clip]{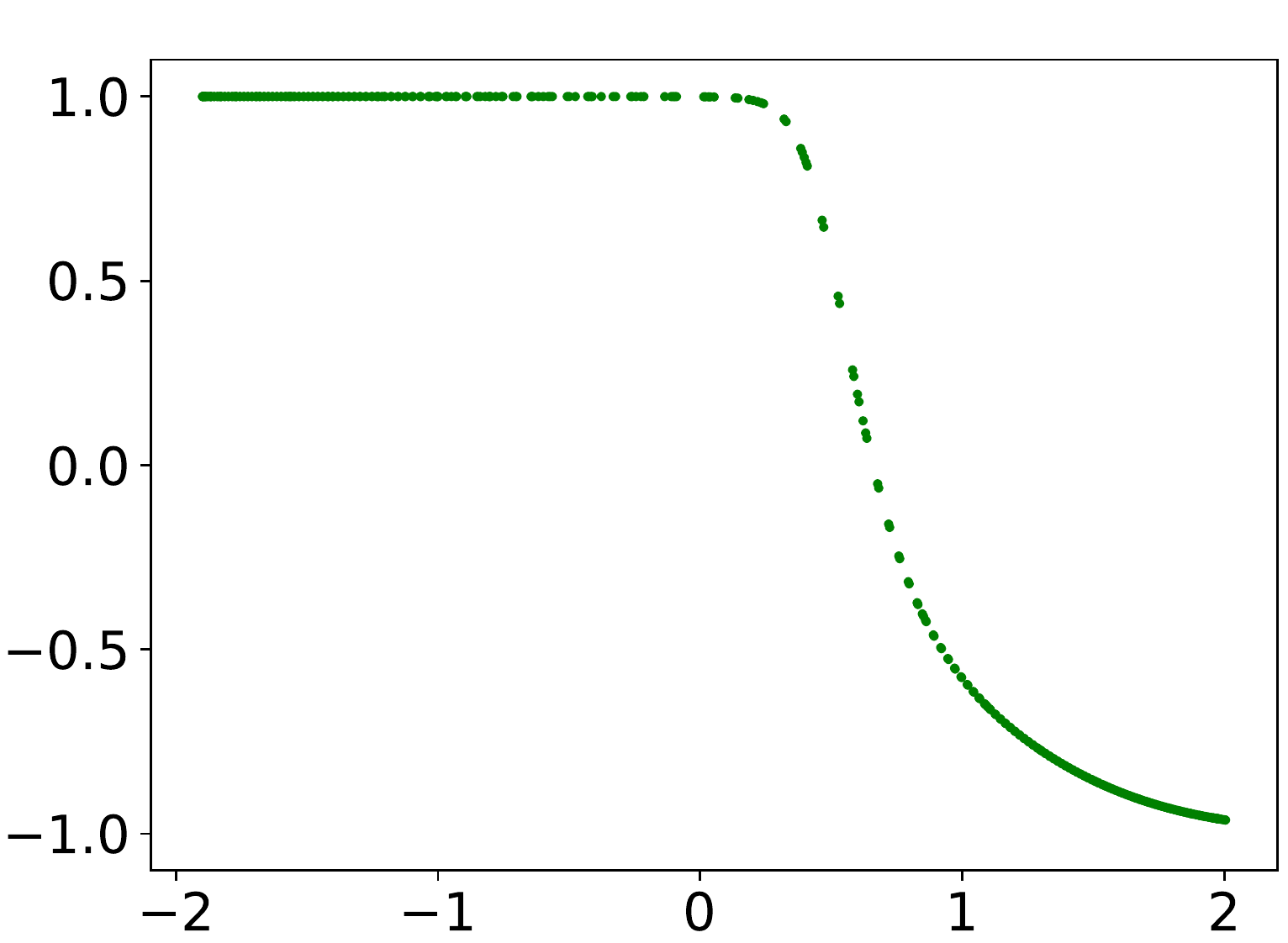}&
	 \includegraphics*[width=0.3\linewidth,trim={0 0 0 4ex},clip]{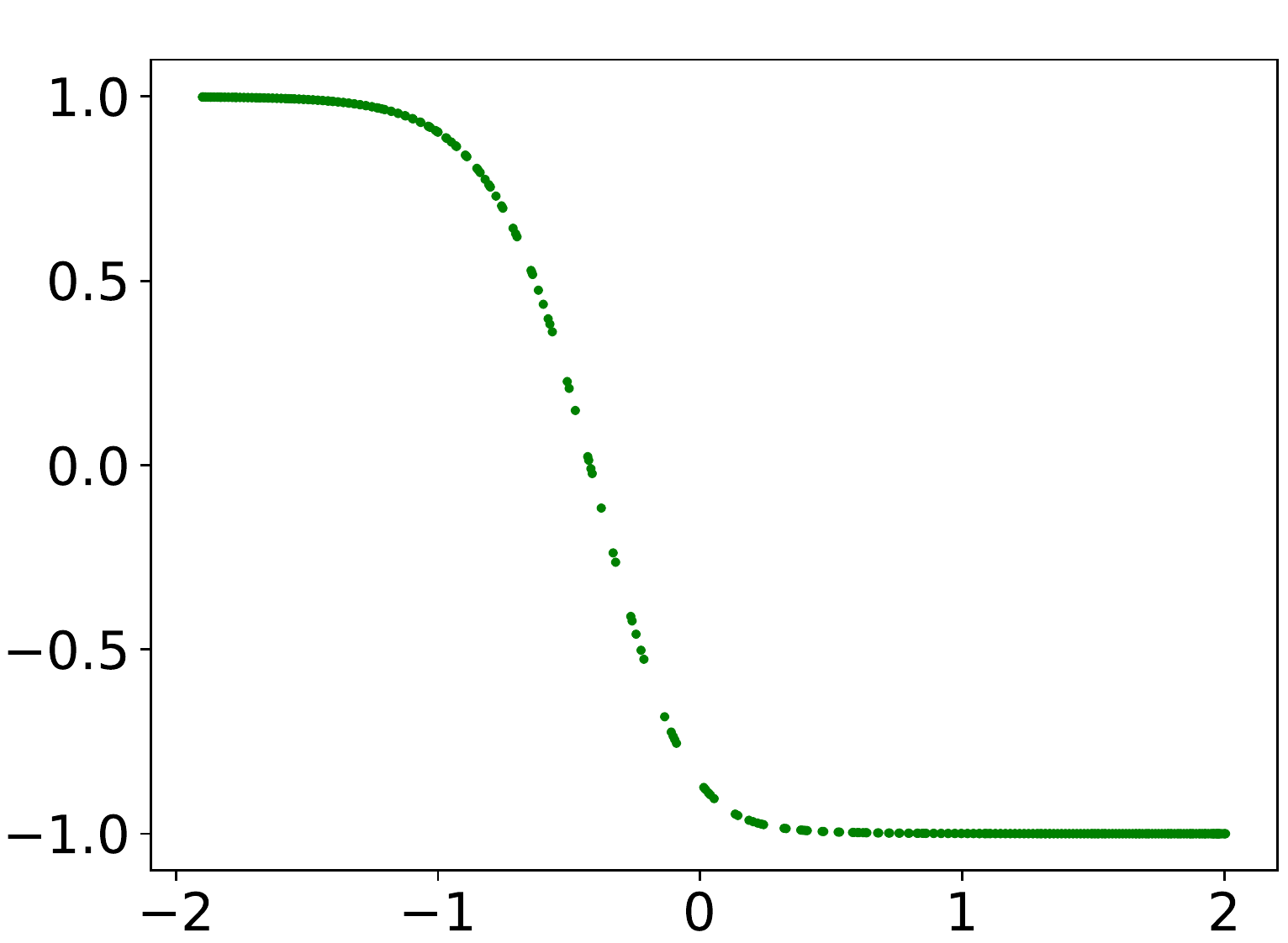}\\[-1.2ex]
	   	& $x_k$ & $x_k$ & $x_k$\\[2ex]
  \end{tabular}
	  \caption{We applied algorithm (\ref{eq:multistepiter}), with neural shape function model for $\widehat{\f}$, to data obtained by simulating the FitzHugh--Nagumo system and then adding $5\%$ noise.  \emph{First panel:} the clean states and predicted states over time. Here the predicted states are obtained by numerically integrating the neural vector field $\widehat{\f}$ forward in time starting at the estimated/filtered initial condition.  \emph{Second panel:} visualization of the learned one-dimensional shape functions $h_{j}(x)$. }
	\label{f:fitz_visual}
  \end{figure}

In Figure \ref{f:fitz_visual}, we plot the predicted and true states for the  neural shape function model of $\widehat{\f}$ learned from data with $5\%$ Gaussian noise.  We also plot the learned one-dimensional neural shape functions $h_j(x)$.  An advantage of this method is the ability to graphically interpret these shape functions, regardless of the dimension $d$ of the vector field being modeled.

Among the methods studied in Figure \ref{f:comp}, only our method and that of \cite{Rudy18a} attempt to deal with noise during training.  Instead of using a proximal term of the form $\| \X - \widehat{\X}^k \|^2$ as in (\ref{eq:multistepX}), \cite{Rudy18a} use a penalty term of the form $\| \X - \Y \|^2$; filtered states are always anchored to the data.  While our method outperforms this penalty-based method, both are more robust to noise than approaches that do not incorporate filtering at all.  Clearly SINDy \cite{Brunton16} has issues with even small amounts of noise, while the dense neural network approach of \cite{Raissi17} encounters problems starting at $5\%$ noise.

We have carried out tests similar to that of Figure \ref{f:comp} for nonlinear, chaotic systems such as the Lorenz, R\"ossler, and double pendulum systems.  For such systems, if we perturb the vector field or initial conditions slightly, sensitive dependence implies that over time, trajectories will diverge exponentially.  For such systems, the prediction error on $[T, 2T]$ can seem large even if we estimate the vector field and filtered states with a reasonable degree of accuracy.  These tests motivate us to measure the \emph{vector field error} $\| \widehat{\f} - \f \|_2$, which provides insight into how well the estimated vector field captures the global behavior of $\f$.

\begin{table}[tbp]
\centering \small
\begin{tabular}{c@{ }c@{ }|c@{ }c@{ }c@{ }l}
\cline{1-5}
Noise &Trajectories & Filter SINDy & \begin{tabular}{@{}c@{}} Neural Shape\\Functions \end{tabular} & Filter DNN &  \\ \cline{1-5}
&1           & $1.77 \times 10^{-4}$        & $0.8467$                 &$0.8157$                &  \\
1 \% &25     & $3.15 \times 10^{-4}$        & $0.2722$                 &$0.3274$                &  \\
&400         & $3.15 \times 10^{-4}$        & $0.2623$                 &$0.2034$                & \\ \cline{1-5}
&1           & $0.2647$         & $0.7773$                 &$0.8667$                &  \\
5 \% &25     & $1.69 \times 10^{-4}$        & $0.4481$                 &$0.2580$                &  \\
&400         & $1.51 \times 10^{-4}$        & $0.2790$                 &$0.2186$                &  \\ \cline{1-5}
&1           & $1.4100$         & $0.8168$                 &$1.7879$                &  \\
10\%&25      & $7.30 \times 10^{-3}$        & $0.4107$                 &$0.2276$                &  \\
&400         & $1.13 \times 10^{-3}$       & $0.3028$                 &$0.2128$                &  \\ 
\end{tabular}
	\caption{For each of three ODE learning methods, all combined with filtering, we record the error in the estimated vector field as a function of both noise strength and the number of trajectories.  For each method and each noise strength, increasing the number of trajectories decreases the vector field error.  Note that retrofitting SINDy with filtering dramatically improves its ability to estimate the vector field accurately, as compared to the results in Figure \ref{f:comp}.}
\label{tab:VF_error}
\end{table}

\emph{We now retrofit both \cite{Raissi17} (the dense neural network model for $\widehat{\f}$) and \cite{Brunton16} (SINDy) with our filtering procedure.}  To be clear, both retrofitted methods follow (\ref{eq:multistepiter}); the only difference is in how $\widehat{\f}$ is modeled.

As all the results shown in Figure \ref{f:comp} are for one trajectory, we now move to the setting where we have multiple trajectories.  We numerically integrate the FitzHugh--Nagumo system with $400$ initial conditions taken from an equispaced $20 \times 20$ grid on the square $[-4,4]$, producing a set of $400$ trajectories.  We then apply each retrofitted procedure to either $1$, $25$ (randomly chosen), or all $400$ trajectories from the collection.  In each case, we obtain an estimated vector field $\widehat{\f}(\x)$.  We compute the vector field error $\| \widehat{\f} - \f \|_2$ using elementary quadrature on the square $[-4,4]$.  Here $\f(\x)$ is the right-hand side of the FitzHugh--Nagumo system; see the ground truth vector field plotted in Figure \ref{fig:fitztruth}.

\begin{figure*}[t] \hspace{-1cm}
	\begin{center}	\begin{tabular}{@{}c@{\hspace{1ex}}c@{}c@{}c@{}c@{}}
		&  Filter SINDy & Neural Shape Functions&  Filter DNN\\
		\hspace{3ex}\rotatebox{90}{\hspace{10ex}{1 Trajectory}} &              
		\includegraphics*[width=0.3\linewidth,trim={2.5cm 2cm 2cm 2.5cm},clip]{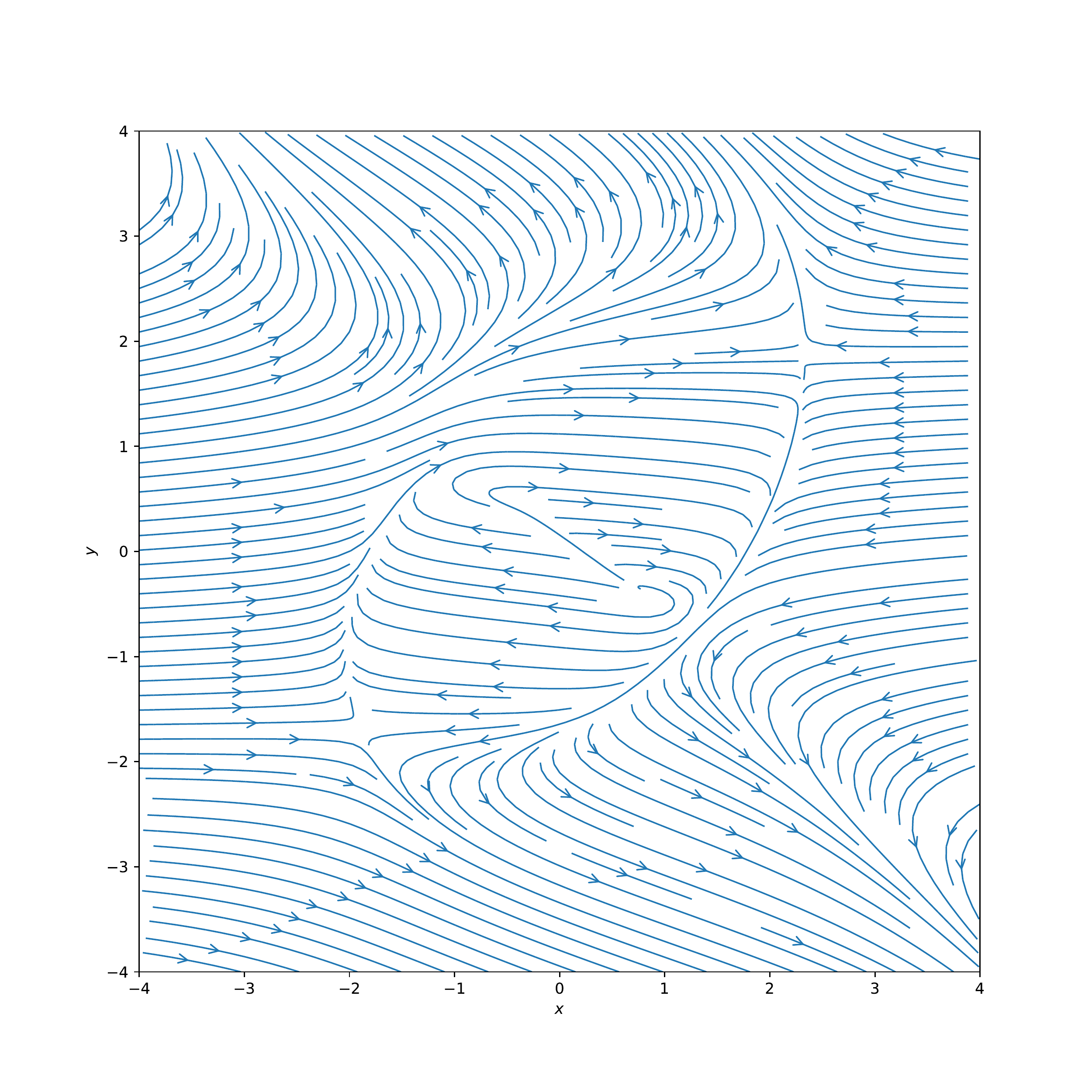}&
		\includegraphics*[width=0.3\linewidth,trim={2.5cm 2cm 2cm 2.5cm},clip]{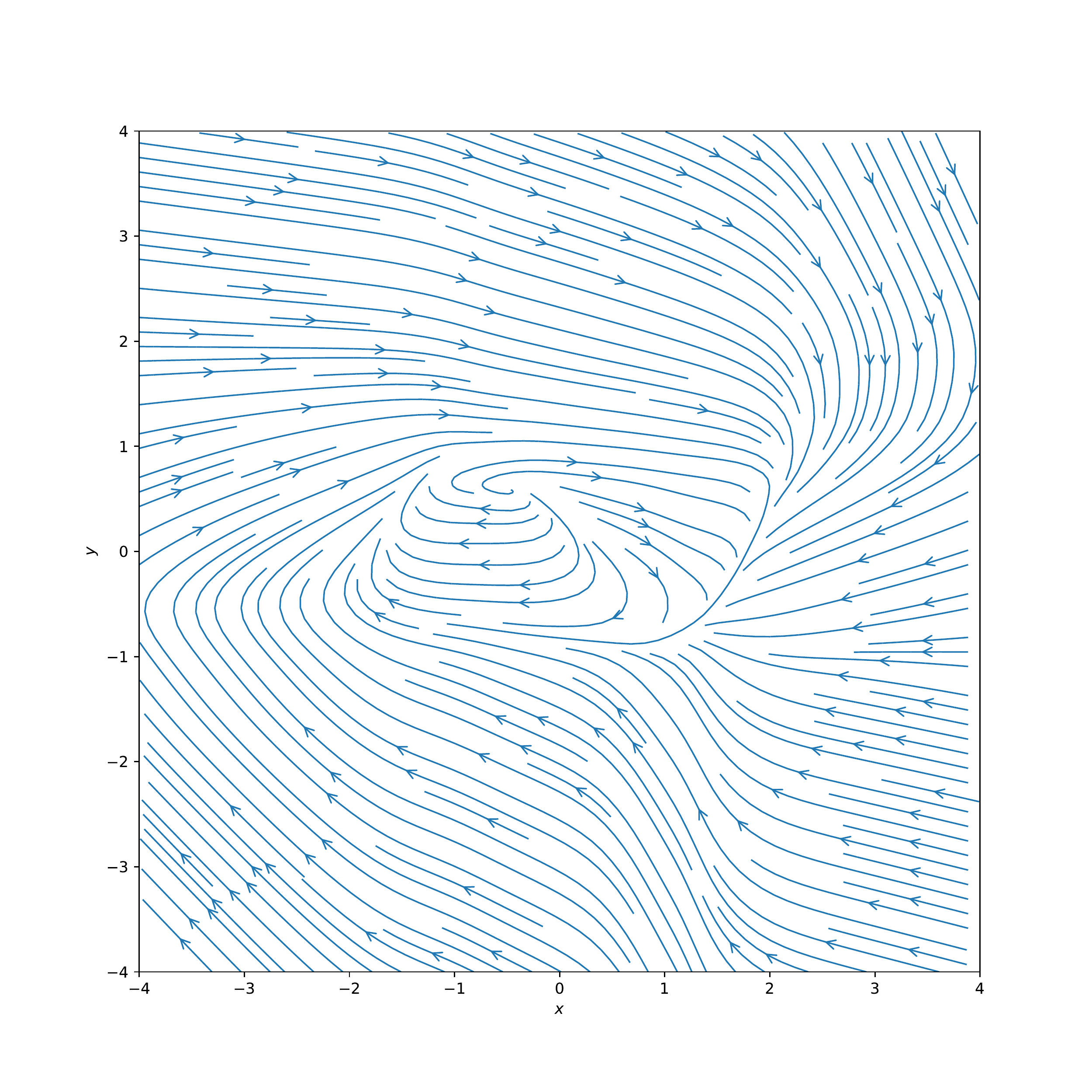}&
		\includegraphics*[width=0.3\linewidth,trim={2.5cm 2cm 2cm 2.5cm},clip]{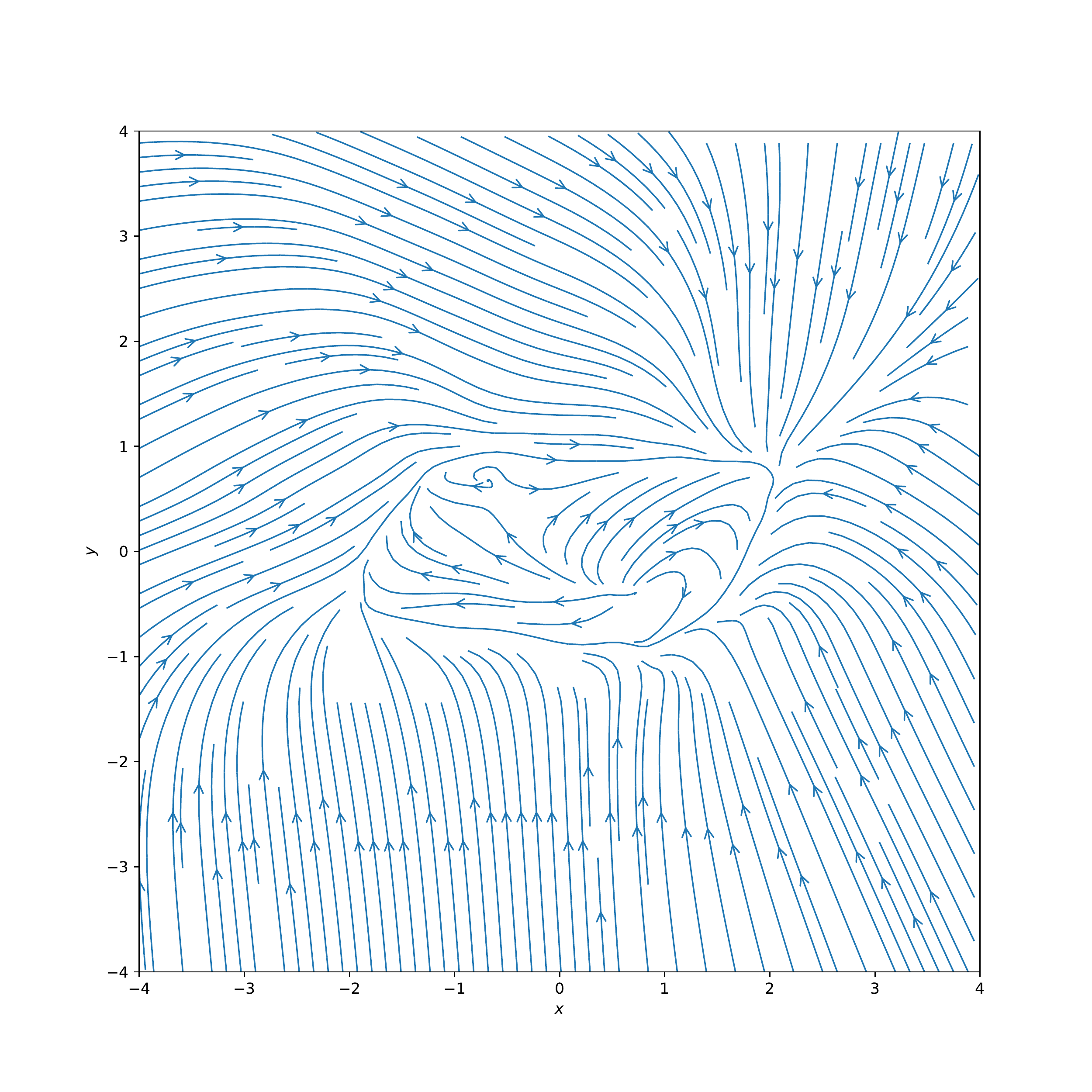}\\
		\hspace{3ex}\rotatebox{90}{\hspace{8ex} {400 Trajectories}}&
		\includegraphics*[width=0.3\linewidth,trim={2.5cm 2cm 2cm 2.5cm},clip]{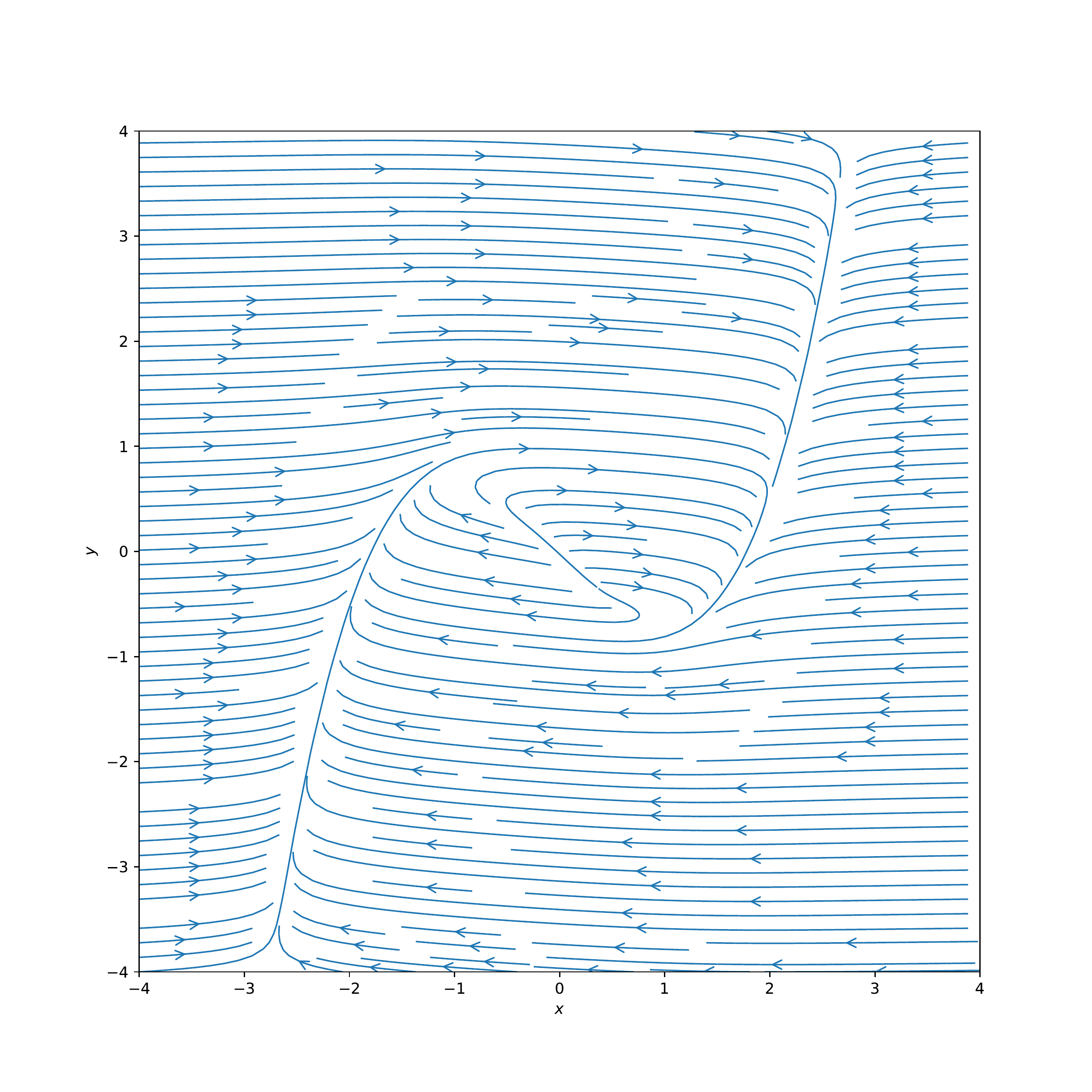}&
		\includegraphics*[width=0.3\linewidth,trim={2.5cm 2cm 2cm 2.5cm},clip]{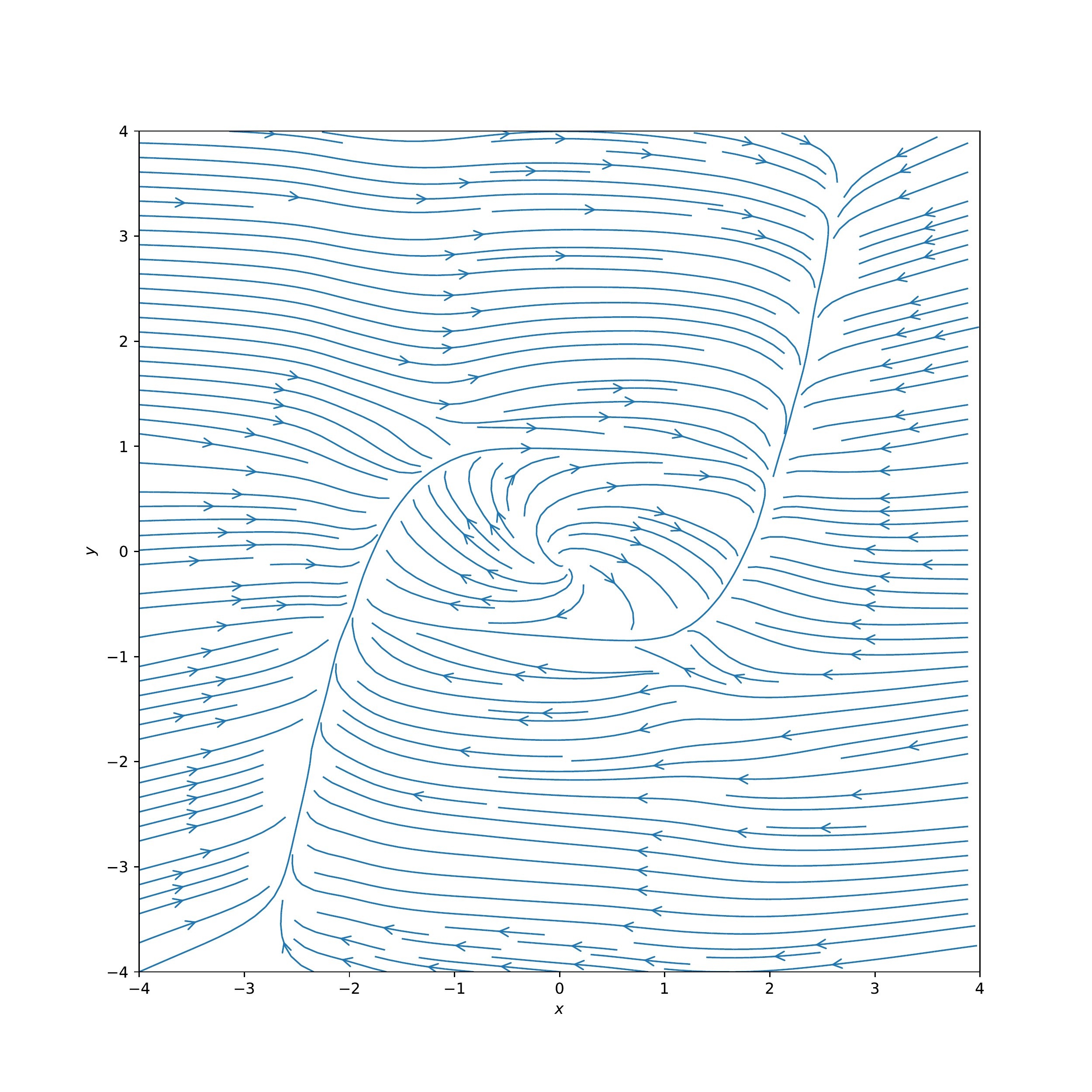}&
		\includegraphics*[width=0.3\linewidth,trim={2.5cm 2cm 2cm 2.5cm},clip]{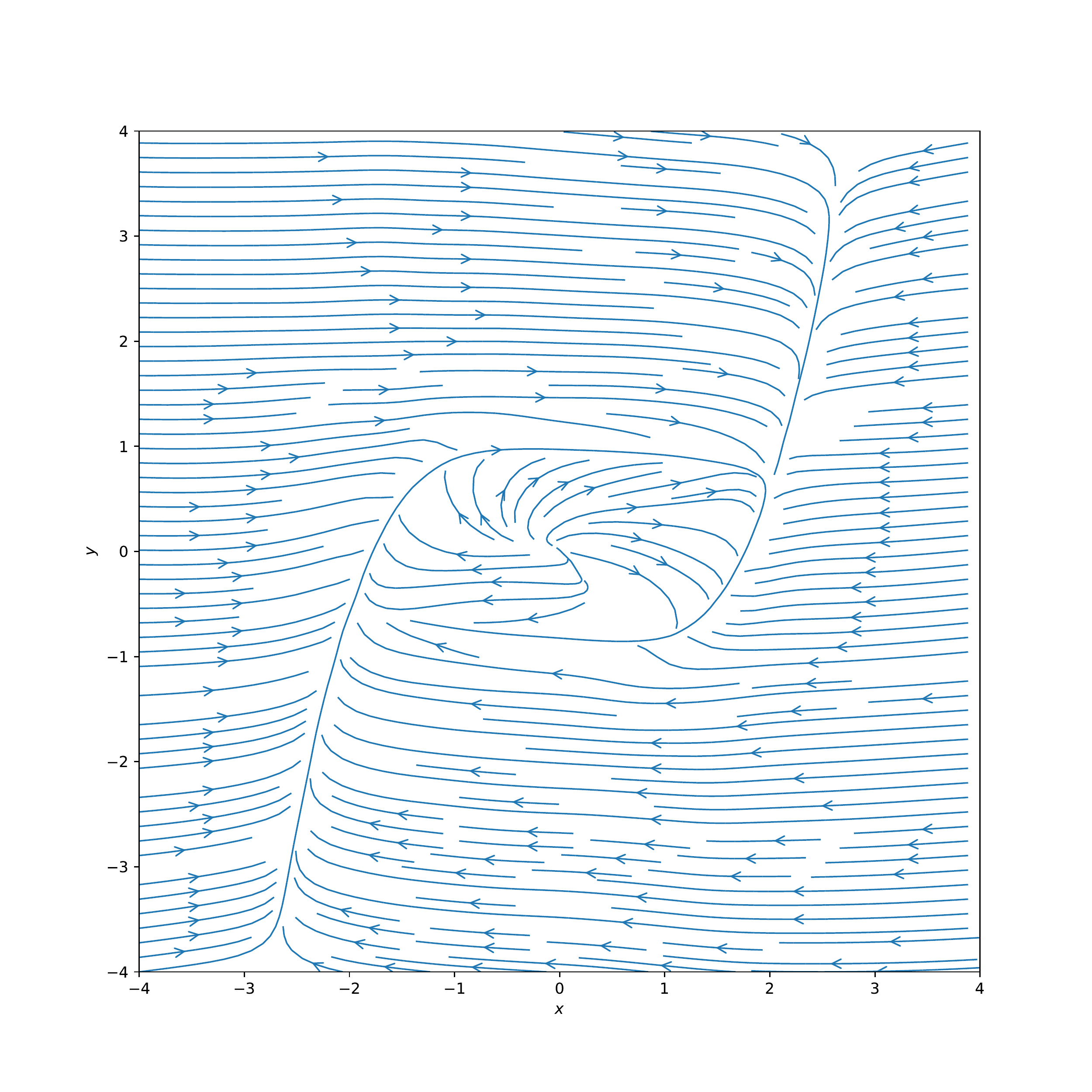}
	\end{tabular} \end{center} \vspace{-0.5cm}
	\caption{We compare the estimated vector fields for data consisting of FitzHugh--Nagumo trajectories corrupted by 10\% Gaussian noise.  We combine the three ODE learning methods from Section \ref{sect:parameterize} with the filtering procedure from Section \ref{sect:opt}.  Note the improvement in the quality of the phase portrait as we increase the number of trajectories used for training.}
	\label{f:pplots}
\end{figure*}

The results, shown in Figure \ref{f:pplots} and Table \ref{tab:VF_error}, clearly show the benefit of combining any method with filtering and increasing the number of trajectories used for training, even if those trajectories are all corrupted with $10\%$ noise.  The improvement is particularly striking for the SINDy method; in fact, after incorporating SINDy into (\ref{eq:multistepiter}), it outperforms the other methods.

\subsection{Nonlinear Oscillator Network}

\begin{table}[tbp]
\centering \small
\begin{tabular}{c@{ }c@{ }|c@{ }c@{ }c@{ }l}
\cline{1-5}
Noise &Trajectories & $\|\f - \widehat{\f}\|$ & $\| \X - \widehat{\X}\|$ & prediction error &  \\ \cline{1-5}
&10           & $5.43 \times 10^{-3}$        & $2.53 \times 10^{-2}$                 &$1.86$                &  \\
1 \% &100     & $5.38 \times 10^{-3}$        & $1.70 \times 10^{-2}$                 &$0.868$                &  \\
&400         & $8.23 \times 10^{-3}$        & $2.83 \times 10^{-2}$                 &$0.975$                & \\ \cline{1-5}
&10           & $8.26 \times 10^{-3}$         & $2.53 \times 10^{-2}$                 &$1.86$                &  \\
5 \% &100     & $8.56 \times 10^{-3}$        & $2.80 \times 10^{-2}$                 &$1.27$                &  \\
&400         & $8.20 \times 10^{-3}$        & $2.81 \times 10^{-2}$                 &$0.984$                &  \\ \cline{1-5}
&10           & $1.77 \times 10^{-1}$         & $6.67 \times 10^{-2}$                 &$3.89$                &  \\
10\%&100      & $1.78 \times 10^{-2}$        & $4.84 \times 10^{-2}$                 &$1.57$                &  \\
&400         & $1.70 \times 10^{-2}$       & $4.69 \times 10^{-2}$                 &$1.17$                &  \\ 
\end{tabular}
	\caption{We apply the retrofitted SINDy algorithm (\ref{eq:bcd}) to noisy observations of the mass-spring system (\ref{eqn:massspring}).  Retrofitting the SINDy algorithm enables it to handle noisy data.  The results also show that, at each noise level, increasing the number of trajectories clearly reduces prediction errors.  For explanations of the different types of errors, please see the main text.}
\label{tab:JAXresults}
\end{table}

Consider a ring of $\mathcal{M}$ masses connected by identical springs, each with potential energy $V(x)$ and force $F(x) = -V'(x)$.  Here $x$ denotes displacement from equilibrium.  Then the equations of motion for the mass-spring system are
\begin{equation}
\label{eqn:massspring}
\ddot{x}_i = F(x_i - x_{i-1}) - F(x_{i+1} - x_i)
\end{equation}
for $i = 1, \ldots, \mathcal{M}$, with the understanding that $x_0 \equiv x_\mathcal{M}$ and $x_1 \equiv x_{\mathcal{M}+1}$.  For our tests, we choose the double-well potential
$V(x) = -8 x^2 + (1/4) x^4$,
so that $F(x) = 16 x - x^3$.

We focus on the system with $\mathcal{M}=3$ masses; when we write the system in first-order form, the system has $2\mathcal{M}=6$ degrees of freedom corresponding to $x_i$ and $\dot{x}_i$ for each $i=1, \ldots, \mathcal{M}$.  To generate data for our tests, we simulate (\ref{eqn:massspring}) using the 8th-order Dormand-Prince integrator in scipy.integrate, with absolute and relative tolerances tuned to $10^{-14}$.  We generate $400$ trajectories, each with an initial condition sampled uniformly from the hypercube $[-1/2, 1/2]^6$.  Each trajectory is saved at $401$ equispaced time steps from $t=0$ to a final time of $t=4$, i.e., $\Delta t = 0.01$.  To these clean trajectories $\X$, we add mean-zero Gaussian noise with strengths of $1\%$, $5\%$, and $10\%$ as before, resulting in noisy data $\Y$.

We focus our attention on a retrofitted SINDy method with $\widehat{\f}$ modeled using a dictionary of polynomials.  In this method, we use a variant of (\ref{eq:multistepiter}) theoretically analyzed in our previous work \cite{RaziBhatICML2019}.  In this method, we incorporate the SINDy model (\ref{eqn:sindy}), which has the benefit of being linear in the parameters $\btheta$.  This linearity implies that the objective function (\ref{eq:objx}) is convex in $\btheta$.  Suppose we split the decision variables $\widehat{\X}$ into two halves, the first half $\widehat{\X}_{+}$ consisting of time steps $j=1, \ldots, \lfloor \mathcal{M}/2 \rfloor$, and the second half $\widehat{\X}_{-}$ consisting of time steps $j=\lfloor \mathcal{M}/2 \rfloor +1, \ldots, \mathcal{M}$.  Then $\widehat{\X} = (\widehat{\X}_{+}, \widehat{\X}_{-})$.  This leads us to the following block coordinate descent algorithm:
\begin{subequations}
	\label{eq:bcd}
	\begin{align}
	\label{eq:bcdtheta}
		\text{train: } \widehat{\btheta}^{k+1} &= \argmin_{\btheta} E(\widehat{\X}_{+}^{k},\widehat{\X}_{-}^{k},\btheta) \\
	\label{eq:bcdxminus}
		\text{filter: } \widehat{\X}_{-}^{k+1} &= \argmin_{\X_{-}} \left\{ \widehat{\X}_{+}^{k},\X_{-},\widehat{\btheta}^{k+1}) + \lambda \| \X_{-} - \widehat{\X}_{-}^{k} \|^2 \right\} \\
	\label{eq:bcdxplus}
		\text{filter: } \widehat{\X}_{+}^{k+1} &= \argmin_{\X_{+}} \left\{ \X_{+},\widehat{\X}_{-}^{k+1},\widehat{\btheta}^{k+1}) + \lambda \| \X_{+} - \widehat{\X}_{+}^{k} \|^2 \right\}.
	\end{align}
\end{subequations}
Splitting $\widehat{\X}$ and formulating the algorithm in this way gives us block convexity.  That is, when we hold two of the three variables in $\{\widehat{\X}_{+}^{k},\widehat{\X}_{-}^{k},\widehat{\btheta}\}$ fixed and minimize over the remaining variable, we obtain in each case a convex subproblem \cite{RaziBhatICML2019}.  Note that this property would not hold if we were to instead use either of the neural network approaches to model $\widehat{\f}$, as in this case (\ref{eq:bcdtheta}) would be non-convex.

\begin{figure}[t!]
\begin{center}\includegraphics[width=3.5in]{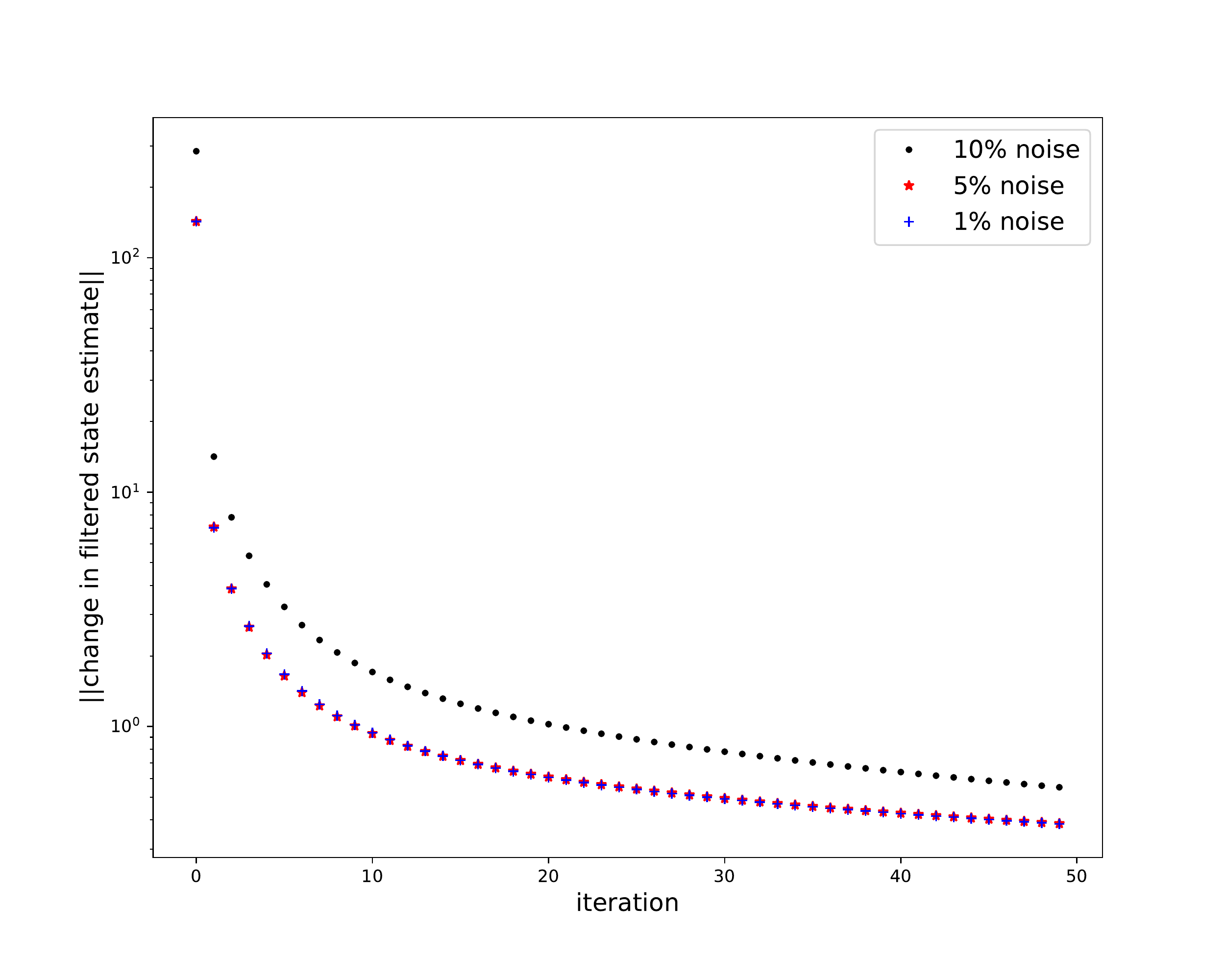}\end{center} \vspace{-0.5cm}
\caption{We monitor and plot the magnitudes of the change in filtered states, $\| \widehat{\X}^{k+1} - \widehat{\X}^k \|$, as a function of iteration number $k$.  We have done this for each of three training sets, each containing the indicated level of noise.  In all cases, we find that this quantity decreases rapidly and monotonically.}
\label{fig:bcd}
\end{figure}

When we train, we restrict $\Y$ to consist of only the first $301$ steps of each trajectory.  Starting with noisy data $\widehat{\X}^0 = \Y$, we run algorithm (\ref{eq:bcd})  with $\lambda = 10^{-8}$.  Note that $\lambda$ multiplies the squared Frobenius error between two large matrices; hence this value of $\lambda$ is still consequential.  We terminate if the norm difference between $\widehat{\X}^{k+1}$ and $\widehat{\X}^k$ is less than $10^{-4}$, or if we have already completed $50$ iterations.

To solve (\ref{eq:bcdxminus}) and (\ref{eq:bcdxplus}), we use $1000$ steps of gradient descent with a learning rate of $3 \times 10^{-2}$.  To solve (\ref{eq:bcdtheta}), we use the iteratively thresholded least squares procedure from \cite{Brunton16, ZhangSchaeffer2019} with a threshold of $0.4$.  For the dictionary $\Xi$ defined in (\ref{eqn:sindy}), we include all polynomials up to degree $3$ in $6$ variables, not including an intercept or constant term, resulting in $s=63$ columns.  We have uploaded our code to \url{https://github.com/hbhat4000/filtersindy/}

\begin{figure}[t!]
\includegraphics[width=3.75in,trim=0 65 0 50,clip]{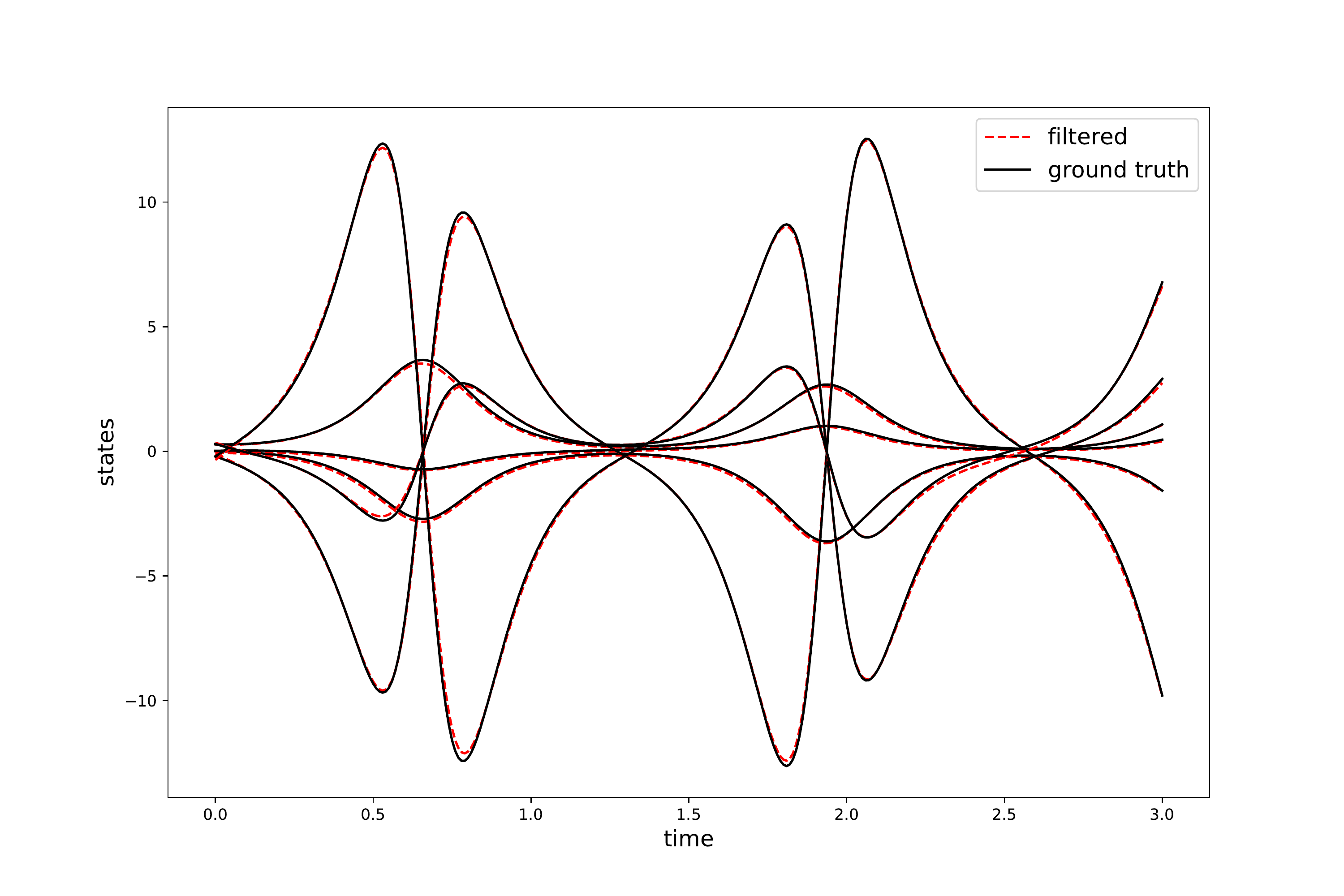} 
\includegraphics[width=3.75in,trim=0 65 0 50,clip]{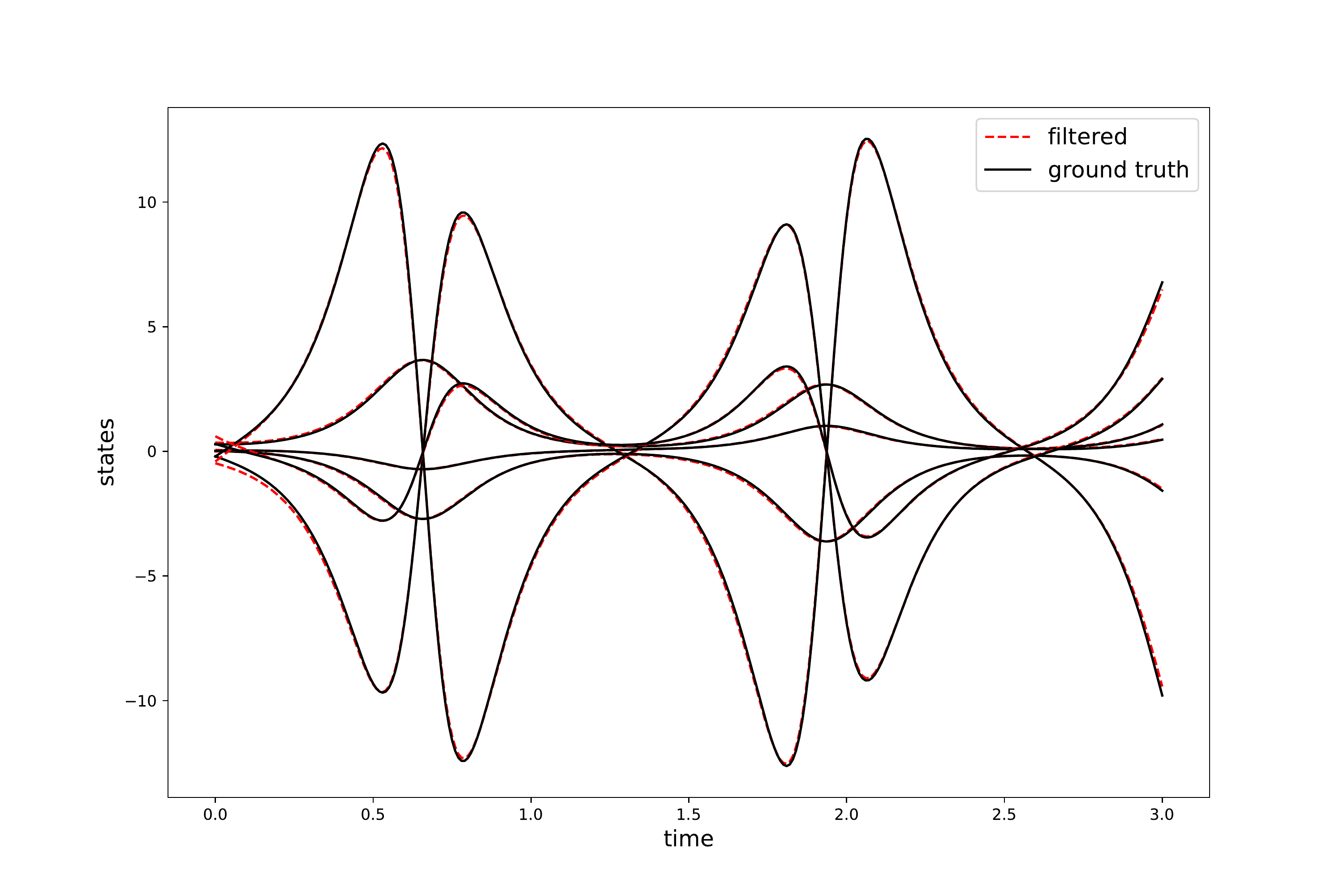} 
\includegraphics[width=3.75in,trim=0 0 0 50,clip]{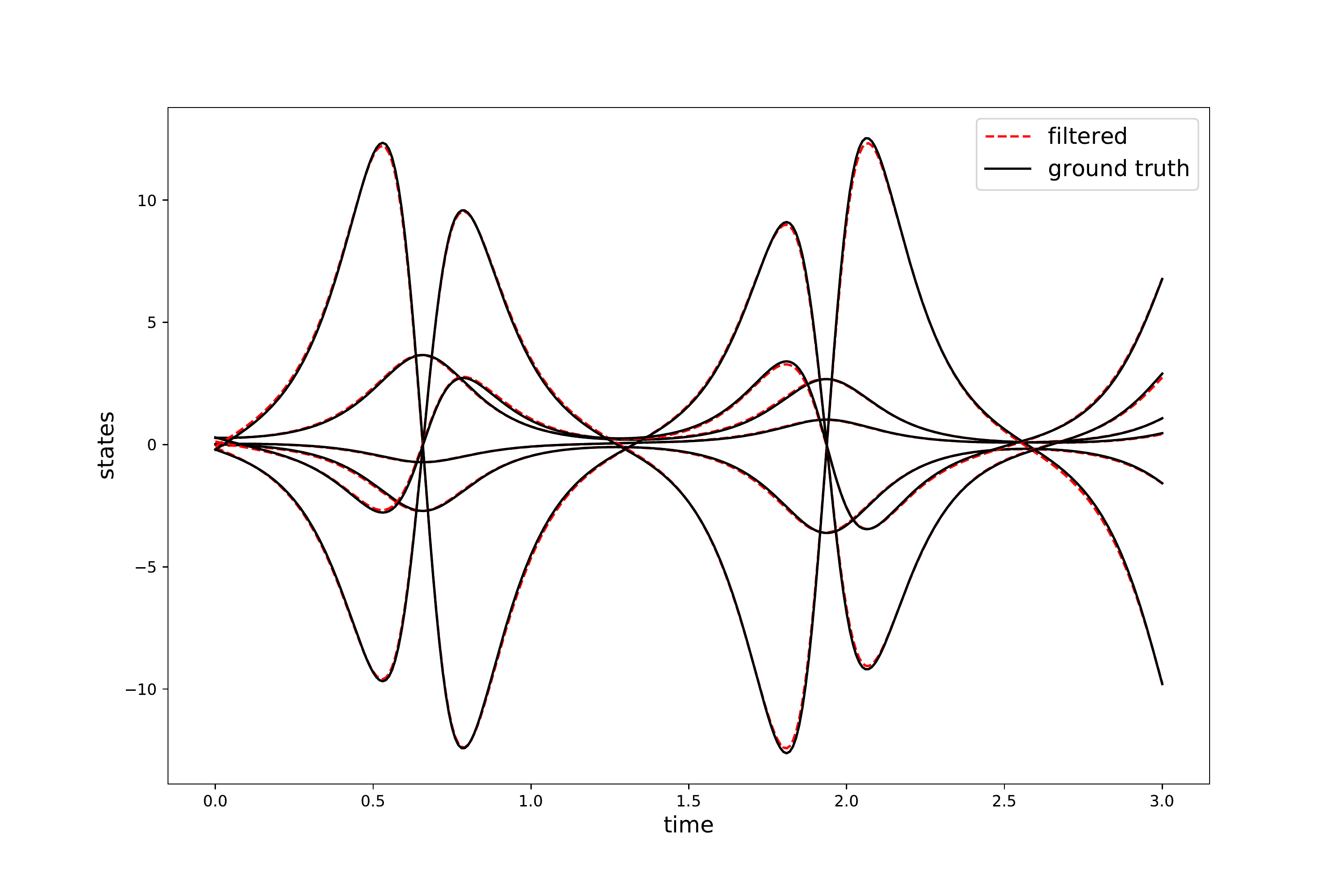}
\caption{We present the results of applying (\ref{eq:bcd}) to $10$ (top), $100$ (middle), or $400$ (bottom) trajectories worth of data for the nonlinear mass-spring system (\ref{eqn:massspring}), all contaminated with $10\%$ Gaussian noise.  In each of these training sets, we have singled out one common trajectory for the purposes of illustration.  In each plot, for this common trajectory, we have plotted the final estimate of the filtered states $\widehat{\X}$ in red, and ground truth trajectories $\X$ in black.  A close inspection of the plots reveals errors with $10$ trajectories; for the purposes of plotting, these errors disappear with $400$ trajectories.  For more details, see the main text.}
\label{fig:filteredtrajectories}
\end{figure}

An interesting property of algorithm (\ref{eq:bcd}) is the way in which $\| \widehat{\X}^{k+1} - \widehat{\X}^k \|$ behaves as a function of $k$.  In Figure \ref{fig:bcd}, we have plotted this quantity as a function of iteration $k$.  We have recorded these magnitudes of the change in filtered states while running algorithm (\ref{eq:bcd}) on the full $400$-trajectory training set, contaminated with either $1\%$, $5\%$, or $10\%$ noise.  In all cases, we see monotonic convergence.  Note that the vertical scale on this plot is logarithmic, further indicating the rapid decrease in  $\| \widehat{\X}^{k+1} - \widehat{\X}^k \|$ as a function of $k$.  As a practical matter, this means that we do not need to run (\ref{eq:bcd}) for many iterations, and that the results are robust to the number of iterations.

After running algorithm (\ref{eq:bcd}) on either $10$, $100$, or $400$ trajectories with either $1\%$, $5\%$, or $10\%$ noise, we quantify errors in three ways---see Table \ref{tab:JAXresults}.  The vector field error $\| \f - \widehat{\f} \|$ measures the mean absolute error between ground truth and estimated vector fields.  The quantity $\| \X - \widehat{\X}\|$ is the mean absolute error between true and filtered states.  After training on the first $301$ steps of each trajectory, we compute predicted trajectories by numerically integrating the estimated vector field forward in time, starting from the filtered initial conditions. The prediction error is the mean absolute error between these predicted trajectories and the true states $\X$.  When we numerically integrate the estimated vector field, we use a standard fourth-order explicit Runge-Kutta method.

The first thing to notice from the results in Table \ref{tab:JAXresults} is that they confirm that combining SINDy with our alternating minimization filtering approach allows SINDy to estimate the mass-spring system's vector field accurately.  Even with 10\% noise in the original data, the vector field error when we train on $400$ trajectories is $1.70 \times 10^{-2}$.  To get a sense for what this number means, note that the ground truth vector field is highly sparse---across the entire ground truth coefficient matrix $\btheta$, out of $63 \cdot 6$ possible entries, only $33$ are nonzero.  The retrofitted SINDy algorithm gets this sparsity pattern $100\%$ correct; if we round the coefficients obtained by SINDy, we obtain the ground truth vector field.

Let us now interpret the filtering errors (or second column) of Table \ref{tab:JAXresults}.  At the 10\% noise level, increasing the number of trajectories appears to decrease the filtering error from $6.67 \times 10^{-2}$ (with 10 trajectories) only slightly to $4.69 \times 10^{-2}$.  To visualize this, we present Figure \ref{fig:filteredtrajectories}.   From top to bottom, we present the results of training with $10$ (top), $100$ (middle), or $400$ (bottom) trajectories worth of data, all contaminated with $10\%$ Gaussian noise.  For the purposes of illustration, we have singled out one trajectory $\{x_i(t), \dot{x}_i(t)\}_{i=1:3}$ that is a member of all three training sets.   Each plot contains $6$ black curves, corresponding to $\{x_i(t), \dot{x}_i(t)\}$ for $i=1, 2, 3$, together with the filtered (or hatted) versions, for a total of $12$ curves per plot.  When there are only $10$ trajectories (top panel), one can discern differences between red and black curves; as we go to $400$ trajectories (bottom panel), these errors mostly disappear.

Overall, Figure \ref{fig:filteredtrajectories} supports the notion that it is indeed possible to estimate filtered states even when the equations of motion (i.e., the vector field) for the underlying system are themselves unknown and must be simultaneously estimated.  However, there is a difference between filtered trajectories and predicted trajectories of the dynamical system.  One will notice from Figure \ref{fig:filteredtrajectories} that the time axis ends at $t=3$; we have used $301$ steps of training data with $\Delta t = 0.01$, so there is no noisy training data to filter beyond $t=3$.

\begin{figure}[t!]
\centering \includegraphics[width=3.75in,trim=0 65 0 50,clip]{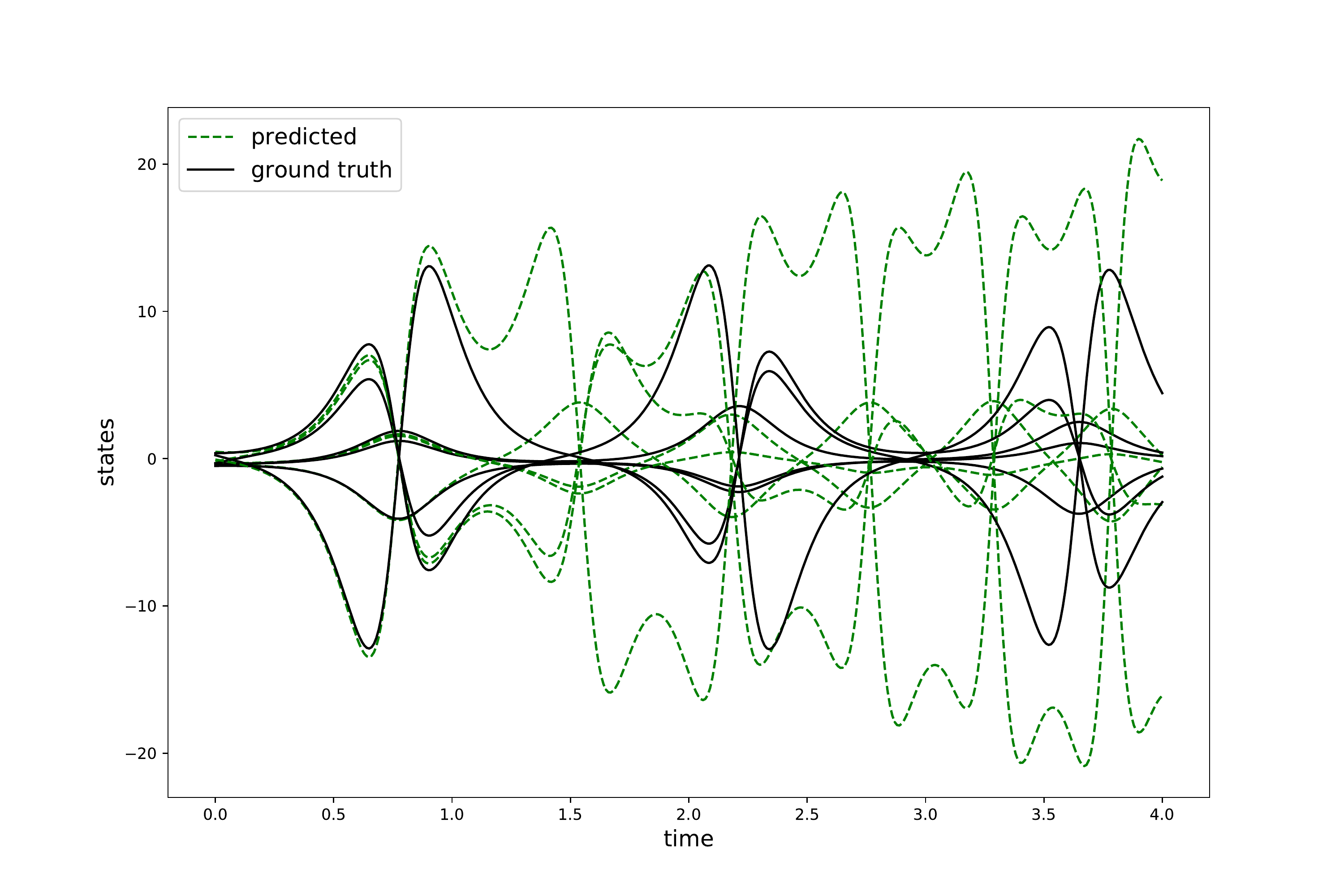} 
\includegraphics[width=3.75in,trim=0 65 0 50,clip]{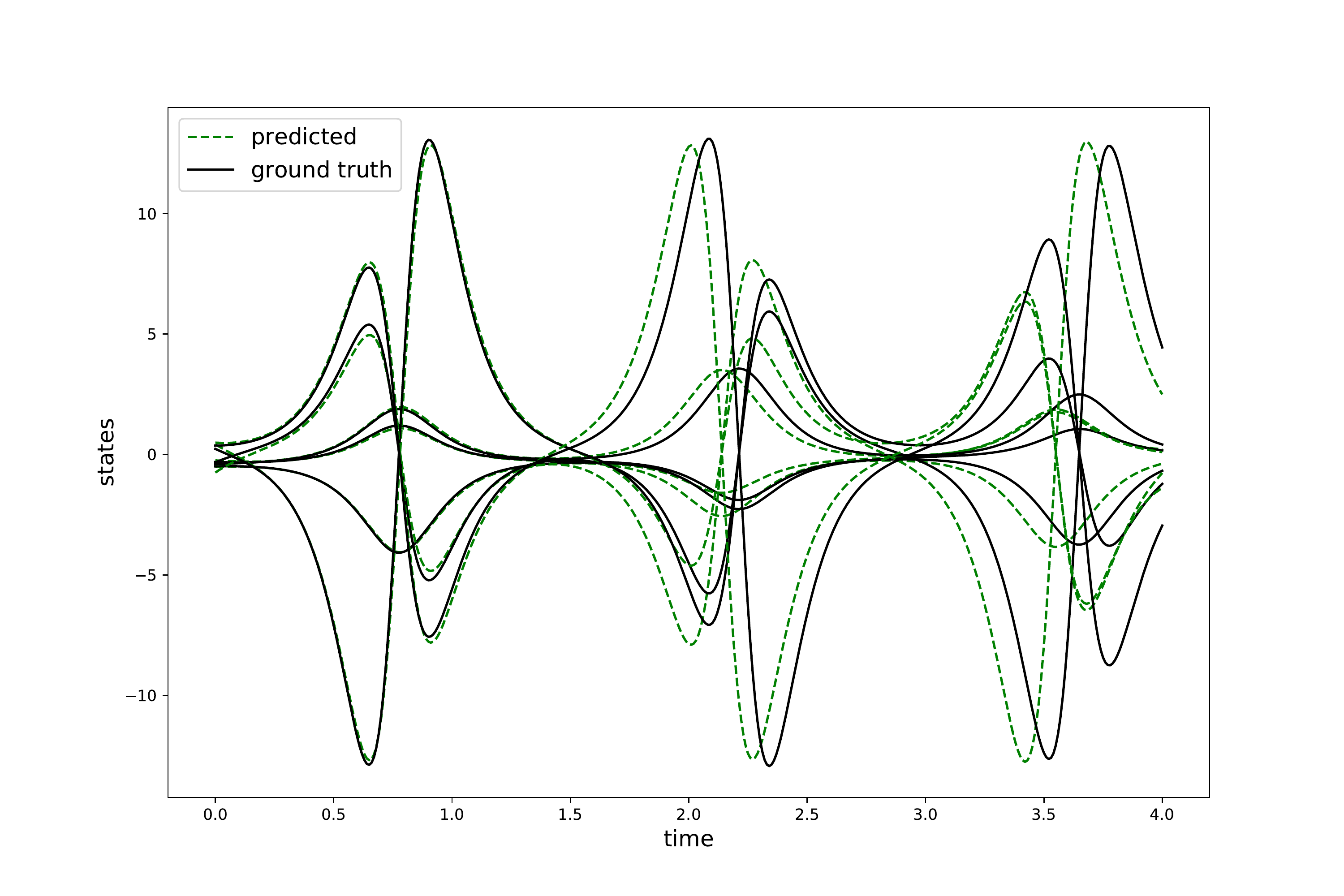} 
\includegraphics[width=3.75in,trim=0 0 0 50,clip]{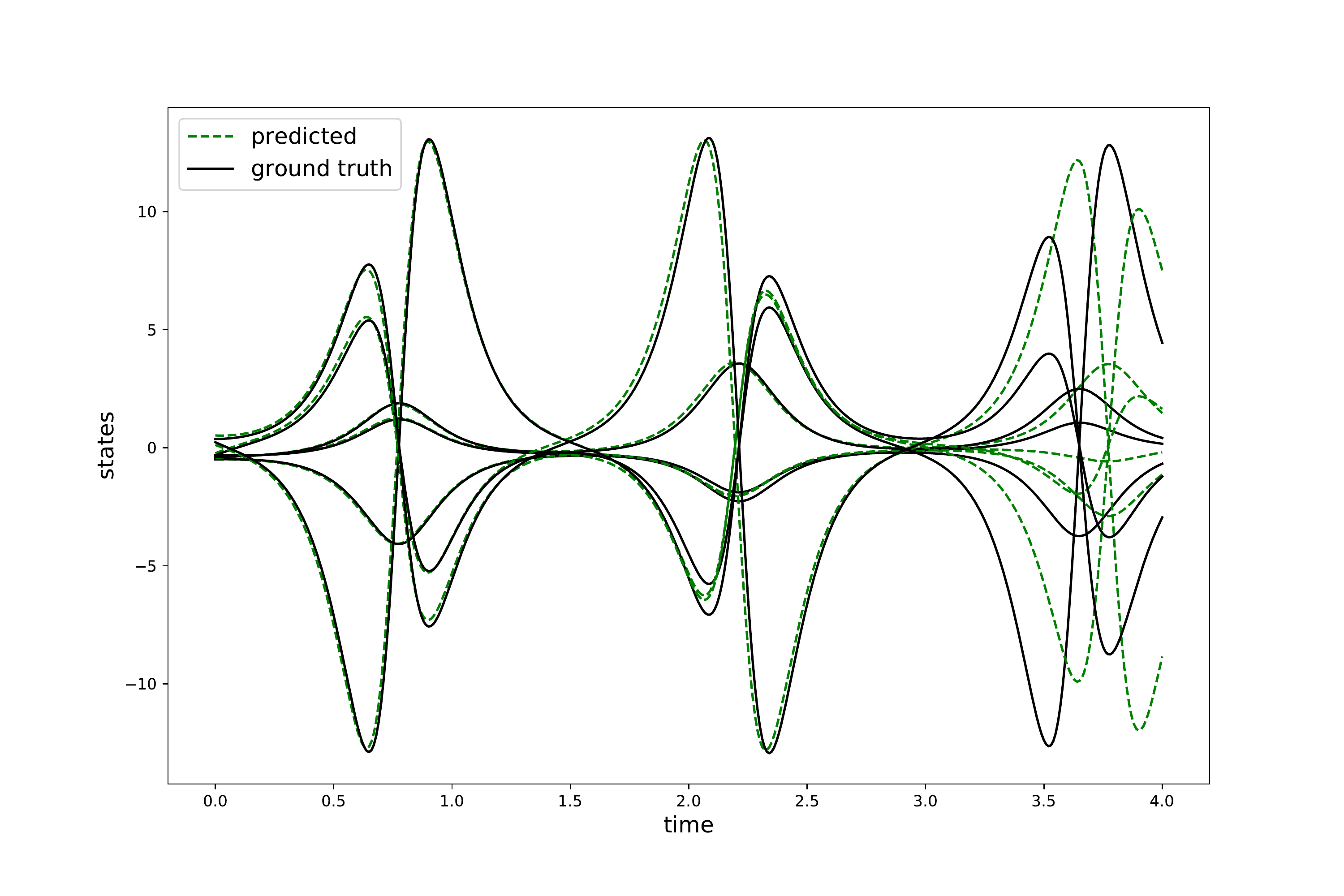}
\caption{We present the results of applying (\ref{eq:bcd}) to $10$ (top), $100$ (middle), or $400$ (bottom) trajectories worth of data from the nonlinear mass-spring system (\ref{eqn:massspring}), all contaminated with $10\%$ Gaussian noise.  In each of these training sets, we have singled out one common trajectory for the purposes of illustration.  In each plot, for this common trajectory, we have plotted predicted trajectories (the results of numerically integrating the estimated vector field forward in time from the estimated initial conditions) in green, and ground truth trajectories $\X$ in black.  The estimated vector field and filtered initial conditions are inaccurate when we use only $10$ trajectories, leading to qualitatively incorrect dynamics.  As we increase the number of trajectories, we recover the qualitatively correct behavior of the mass-spring system.  The dynamics for $t \in [0,3]$ are more accurate because that is the interval covered by the training data; for $t \in (3,4]$, we are seeing the results of propagating beyond the training interval.}
\label{fig:predictedtrajectories}
\end{figure}

To obtain solutions of the estimated dynamical system beyond $t=3$, we must numerically integrate, as in the predicted trajectories whose errors are quantified in the third column of Table \ref{tab:JAXresults}.  When we compute the predicted trajectories in this table, we integrate all the way up to $t=4$ and then compare against the ground truth (clean) states of the system $\X$.  We view predictions on $t \in (3,4]$ as a true test set, i.e., a test of the estimated vector field's ability to extrapolate beyond the training set.

We again single out the same trajectory common to our training sets with $10$, $100$, or $400$ total trajectories.  Starting with the estimated/filtered initial conditions, we numerically integrate the estimated vector field forward in time using a standard fourth-order explicit Runge-Kutta method, from $t=0$ to $t=4$.  In Figure \ref{fig:predictedtrajectories}, we compare the results of these numerical integrations (in green) against the ground truth trajectories (in black).  When the number of trajectories is small (top), the predicted dynamics are qualitatively wrong.  As we train on more trajectories (middle, bottom), the dynamics begin to qualitatively match the true mass-spring system's dynamics.  Note that we obtain much better quantitative accuracy for $t \in [0,3]$, the training interval.

\begin{figure}[t!]
\includegraphics[width=3.75in]{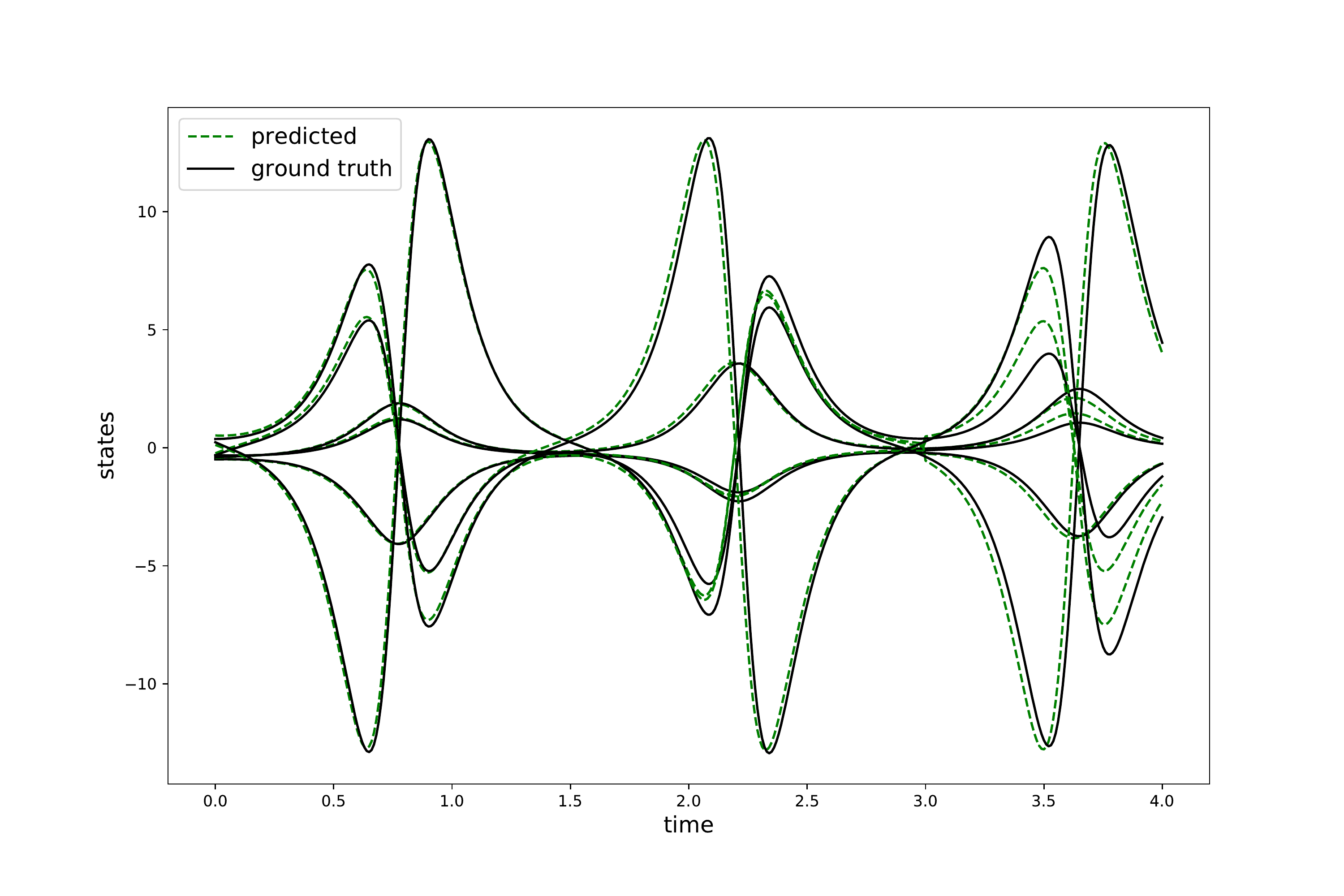}
\caption{We continue with results for algorithm (\ref{eq:bcd}) applied to noisy data from the nonlinear mass-spring system (\ref{eqn:massspring}).  We redo the numerical integration from the bottom panel of Figure \ref{fig:predictedtrajectories}, this time resetting the state of the system at $t=3$ to equal our estimate $\widehat{\x}(3)$ of the true state of the system at that time.  This improves the quantitative accuracy of our predicted trajectory on the extrapolation/test interval $t \in (3,4]$.}
\label{fig:resetting}
\end{figure}

For the bottom panel (trained on $400$ trajectories), by the time we reach $t=3$, the predicted state has drifted noticeably away from the ground truth.  We can improve upon this situation by \emph{resetting} the state of the system, at $t=3$, to equal our estimate $\widehat{\x}(3)$ of the true state at that time and then \emph{continuing} the numerical integration until $t=4$.  In Figure \ref{fig:resetting}, we see that this procedure leads to improved quantitative agreement between predicted and ground truth trajectories on the test interval $t \in (3,4]$.

\subsection{MicroPMU Data}
We apply the neural shape function plus filtering method to $\mu$PMU data taken from a Mountain View substation near Riverside, CA.  We use one hour's worth of data from both Aug. 9 and Aug. 1, 2017.  We use measurements aggregated at a spacing of $\Delta t = 0.01$.  In prior work, these dates have been identified as corresponding to normal (Aug. 9) and anomalous (Aug. 1) system operation.  Our idea is to learn two vector fields, one for normal and one for anomalous behavior.

To begin, we prefiltered the data using a wavelet low-pass filter.  This was to remove high-frequency noise prevalent in the raw data.  Next, we focused our modeling efforts on two phase angle variables, $\theta_1$ and $\theta_2$.  This means that we did not use $10$ components of the full $12$-dimensional signal.  To implement the multiple trajectory idea, we reshaped the single hour's trajectory from its original length of $T = 360000$ to $N = 3600$ trajectories each consisting of $T = 100$ points.  To implement (\ref{eq:multisteptheta}) we use $100$ epochs of the Adam optimizer with a learning rate of $0.002$; to implement (\ref{eq:multistepX}), we use L-BFGS-B from scipy.optimize with default tolerances and $\lambda = 1000$.  The network itself consists of $B = 3$ shape functions with a depth of $D = 2$ and $U = 16$ units per layer.   The results are robust to increasing $\lambda$, $B$, $D$, or $U$.  However, note that training on one prefiltered trajectory of length $T = 360000$, without alternating filtering, does not yield usable models.

\begin{figure}
	\begin{center}
	\includegraphics[width=0.45\linewidth, trim={2.5cm 2cm 2cm 2.5cm}, clip]{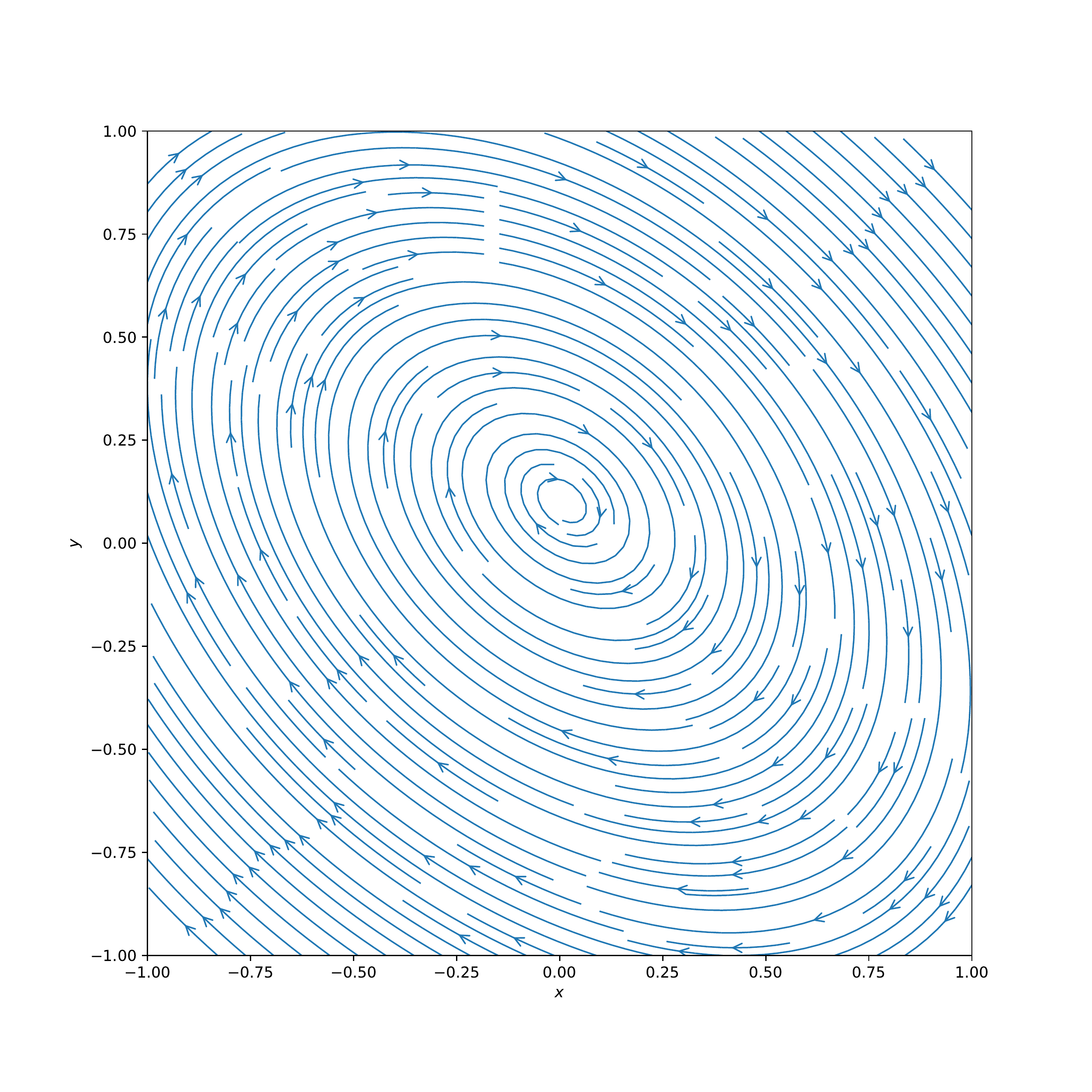}
	\includegraphics[width=0.45\linewidth, trim={2.5cm 2cm 2cm 2.5cm}, clip]{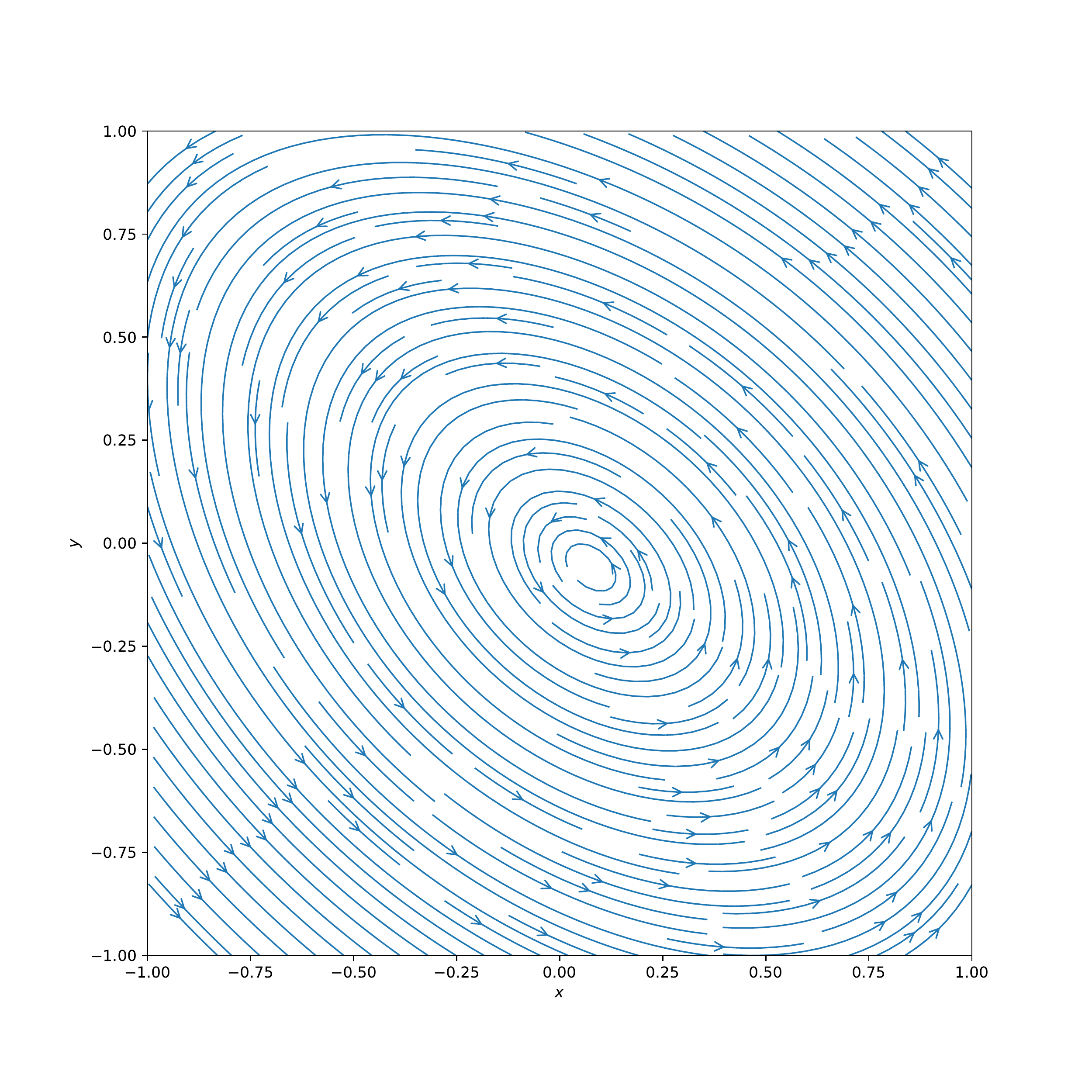}
	\end{center} \vspace{-0.75cm}
	\caption{$\mu$PMU phase plots for normal (left) and anomalous (right) system behavior. Note that the fixed point is stable (eigenvalues $-4.5014 \times 10^{-4} \pm 1.01 i$) on the left and unstable (eigenvalues $4.7166 \times 10^{-5} \pm 1.01i$) on the right.}
	\label{fig:mpmu}
\end{figure}

In Figure \ref{fig:mpmu}, we plot the phase portraits for the learned systems.  Interestingly, the main difference between the learned vector fields for the normal (left) and anomalous (right) settings has to do with stability.  In short, we see that normal (respectively, anomalous) system operation corresponds to stable (respectively, unstable) oscillations.  We have confirmed this by numerically finding the fixed points of the vector fields and checking the eigenvalues of the Jacobians at these fixed points.

\section{Discussion}
We find that we can compensate for noisy training data by increasing its volume (the number of trajectories) and by applying simultaneous filtering.  Filtering, in the form of the proximal step (\ref{eq:multistepX}), reestimates the states $\widehat{\X}$ based on the best model at the current iteration, $\widehat{\f}(\X; \widehat{\btheta}^k)$.  When $\Y$ is heavily contaminated by noise, several iterations of (\ref{eq:multistepX}) allow $\widehat{\X}$ to step away from $\Y$ as needed.  While we have focused on the FitzHugh--Nagumo and nonlinear mass-spring systems in the paper, we are currently running additional tests for higher-dimensional physical systems.  Preliminary results confirm the same trends we have observed for the FitzHugh--Nagumo system.  

SINDy uses polynomial shape functions and the FitzHugh--Nagumo vector field consists of polynomials.  Yet unless we include filtering and multiple trajectories, SINDy performs poorly on noisy data.  For non-polynomial vector fields, the neural network approaches allow for greater model flexibility.  We hypothesize that on non-polynomial systems, neural network-based approaches will prove superior.  With the neural shape function approach, one can plot and visualize the one-dimensional shape functions $h_j : \mathbb{R} \to \mathbb{R}$, analogous to visualizing the components of $\Xi$ in SINDy.  However, due to its special form and construction, the neural shape function model is more difficult to train than a standard dense neural network.  

In future work, we seek to replace these dense neural networks with sparsely connected neural networks that have optimal approximation properties \cite{Bolcskei2019}.  We will also combine these methods with dimensionality reduction techniques to obtain reduced-order models from the full $12$-dimensional $\mu$PMU time series.  Ultimately, our goal is to uncover global phenomena, such as instabilities, attractors, and periodic orbits, that go beyond single trajectory forecasts.

\end{document}